\documentclass{article} 
\usepackage{iclr2025_conference,times}

\usepackage{amsmath,amsfonts,bm}

\def\eqref#1{equation~\ref{#1}}

\def\1{\bm{1}}

\def\rd{{\textnormal{d}}}

\def\rvx{{\mathbf{x}}}
\def\rvy{{\mathbf{y}}}
\def\rvz{{\mathbf{z}}}

\def\vf{{\bm{f}}}

\def\vs{{\bm{s}}}

\def\mF{{\bm{F}}}

\def\mI{{\bm{I}}}

\DeclareMathAlphabet{\mathsfit}{\encodingdefault}{\sfdefault}{m}{sl}
\SetMathAlphabet{\mathsfit}{bold}{\encodingdefault}{\sfdefault}{bx}{n}

\def\gL{{\mathcal{L}}}

\def\gN{{\mathcal{N}}}

\newcommand{\pdata}{p_{\rm{data}}}

\newcommand{\E}{\mathbb{E}}

\usepackage{hyperref}
\usepackage{url}
\usepackage{cleveref}
\usepackage{booktabs}
\usepackage{multirow}
\usepackage{subcaption}
\usepackage{adjustbox} 
\usepackage{float}

\usepackage{multirow}

\usepackage{graphicx}
\title{Improved Training Technique for Latent Consistency Models}

\iclrfinalcopy

\author{Quan Dao$^{*\dagger}$\\
Rutgers University \\
\texttt{quan.dao@rutgers.edu} \\ \And 
Khanh Doan$^{*}$\\
Movian AI, Vietnam \\
\texttt{dnkhanh.k63.bk@gmail.com} \\ \And
Di Liu\\
Rutgers University \\
\texttt{di.liu@rutgers.edu} \\   \And
Trung Le\\
Monash University \\
\texttt{trunglm@monash.edu} \\   \And
Dimitris Metaxas\\
Rutgers University \\
\texttt{dnm@cs.rutgers.edu} \\
}

\newcommand{\minisection}[1]{\noindent{\textbf{#1}}}

\begin{document}

\maketitle
\def\thefootnote{\textsuperscript{$*$}}\footnotetext{Equal contributions.}
\def\thefootnote{\textsuperscript{$\dagger$}}\footnotetext{Project Lead \& Corresponding Author.}

\begin{abstract}
Consistency models are a new family of generative models capable of producing high-quality samples in either a single step or multiple steps. Recently, consistency models have demonstrated impressive performance, achieving results on par with diffusion models in the pixel space. However, the success of scaling consistency training to large-scale datasets, particularly for text-to-image and video generation tasks, is determined by performance in the latent space. In this work, we analyze the statistical differences between pixel and latent spaces, discovering that latent data often contains highly impulsive outliers, which significantly degrade the performance of iCT in the latent space. To address this, we replace Pseudo-Huber losses with Cauchy losses, effectively mitigating the impact of outliers. Additionally, we introduce a diffusion loss at early timesteps and employ optimal transport (OT) coupling to further enhance performance. Lastly, we introduce the adaptive scaling-$c$ scheduler to manage the robust training process and adopt Non-scaling LayerNorm in the architecture to better capture the statistics of the features and reduce outlier impact. With these strategies, we successfully train latent consistency models capable of high-quality sampling with one or two steps, significantly narrowing the performance gap between latent consistency and diffusion models. The implementation is released here: \url{https://github.com/quandao10/sLCT/}
\end{abstract}

\section{Introduction}
In recent years, generative models have gained significant prominence, with models like ChatGPT excelling in language generation and Stable Diffusion \citep{rombach2021highresolution}. In computer vision, the diffusion model \citep{song2020score, song2019generative, ho2020denoising, sohl2015deep} has quickly popularized and dominated the Adversarial Generative Model (GAN) \citep{goodfellow2014generative}. It is capable of generating high-quality diverse images that beat SoTA GAN models \citep{dhariwal2021diffusion}. Additionally, diffusion models are easier to train, as they avoid the common pitfalls of training instability and the need for meticulous hyperparameter tuning associated with GANs. The application of diffusion spans the entire computer vision field, including text-to-image generation \citep{rombach2021highresolution, gu2022vector}, image editing \citep{meng2021sdedit, cyclediffusion, huberman2024edit, han2024proxedit, he2024dice}, text-to-3D generation \citep{poole2022dreamfusion, wang2024prolificdreamer}, personalization \citep{ruiz2022dreambooth, van2023anti, kumari2023multi} and control generation \citep{zhang2023adding, brooks2022instructpix2pix, zhangli2024layout}. Despite their powerful capabilities, they require thousands of function evaluations for sampling, which is computationally expensive and hinders their application in the real world. Numerous efforts have been made to address this sampling challenge, either by proposing new training frameworks \citep{xiao2021tackling, rombach2021highresolution} or through distillation techniques \citep{meng2023distillation, yin2024one, sauer2023adversarial, dao2024self}. However, methods like \citep{xiao2021tackling} suffer from low recall due to the inherent challenges of GAN training, while \citep{rombach2021highresolution} still requires multi-step sampling. Distillation-based approaches, on the other hand, rely heavily on pretrained diffusion models and demand additional training.

Recently, \citep{song2023consistency} introduced a new family of generative models called the consistency model. Compared to the diffusion model \citep{song2019generative, song2020score, ho2020denoising}, the consistency model could both generate high-quality samples in a single step and multi-steps. The consistency model could be obtained by either consistency distillation (CD) or consistency training (CT). In previous work \citep{song2023consistency}, CD significantly outperforms CT. However, the CD requires additional training budget for using pretrained diffusion, and its generation quality is inherently limited by the pretrained diffusion. Subsequent research \citep{song2023improved} improves the consistency training procedure, resulting in performance that not only surpasses consistency distillation but also approaches SoTA performance of diffusion models. Additionally, several works \citep{kim2023consistency, geng2024consistency} have further enhanced the efficiency and performance of CT, achieving significant results. However, all of these efforts have focused exclusively on pixel space, where data is perfectly bounded. In contrast, most large-scale applications of diffusion models, such as text-to-image or video generation, operate in latent space \citep{rombach2021highresolution, gu2022vector}, as training on pixel space for large-scale datasets is impractical. Therefore, to scale consistency models for large datasets, the consistency must perform effectively in latent space. This work addresses the key question: How well can consistency models perform in latent space? To explore this, we first directly applied the SoTA pixel consistency training method, iCT \citep{song2023improved}, to latent space. The preliminary results were extremely poor, as illustrated in \cref{fig:qualitative_ict}, motivating a deeper investigation into the underlying causes of this suboptimal performance. We aim to improve CT in latent space, narrowing the gap between the performance of latent consistency and diffusion.

We first conducted a statistical analysis of both latent and pixel spaces. Our analysis revealed that the latent space contains impulsive outliers, which, while accounting for a very small proportion, exhibit extremely high values akin to salt-and-pepper noise. We also drew a parallel between Deep Q-Networks (DQN) and the Consistency Model, as both employ temporal difference (TD) loss. This could lead to training instability compared to the Kullback-Leibler (KL) loss used in diffusion models. Even in bounded pixel space, the TD loss still contains impulsive outliers, which \citep{song2023improved} addressed by proposing the use of Pseudo-Huber loss to reduce training instability. As shown in \cref{fig:impulsive_noise}, the latent input contains extremely high impulsive outliers, leading to very large TD values. Consequently, the Pseudo-Huber loss fails to sufficiently mitigate these outliers, resulting in poor performance as demonstrated in \cref{fig:qualitative_ict}. To overcome this challenge, we adopt Cauchy loss, which heavily penalizes extremely impulsive outliers. Additionally, we introduce diffusion loss at early timesteps along with optimal transport (OT) matching, both of which significantly enhance the model's performance. Finally, we propose an adaptive scaling $c$ schedule to effectively control the robustness of the model, and we incorporate Non-scaling LayerNorm into the architecture. With these techniques, we significantly boost the performance of latent consistency model compared to the baseline iCT framework and bridge the gap between the latent diffusion and consistency training.

\section{Related Works} \label{related}

Consistency model \citep{song2023consistency, song2023improved} proposes a new type of generative model based on PF-ODE, which allows 1, 2 or multi-step sampling. The consistency model could be obtained by either training from scratch using an unbiased score estimator or distilling from a pretrained diffusion model. Several works improve the training of the consistency model. ACT \citep{kong2023act}, CTM \citep{kim2023consistency} propose to use additional GAN along with consistency objective. While these methods could improve the performance of consistency training, they require an additional discriminator, which could need to tune the hyperparameters carefully. MCM \citep{heek2024multistep} introduces multistep consistency training, which is a combination of TRACT \citep{berthelot2023tract} and CM \citep{song2023consistency}. MCM increases the sampling budget to 2-8 steps to tradeoff with efficient training and high-quality image generation. ECM \citep{geng2024consistency} initializes the consistency model by pretrained diffusion model and fine-tuning it using the consistency training objective. ECM vastly achieves improved training times while maintaining good generation performance. However, ECM requires pretrained diffusion model, which must use the same architecture as the pretrained diffusion architecture. Although these works successfully improve the performance and efficiency of consistency training, they only investigate consistency training on pixel space. As in the diffusion model, where most applications are now based on latent space, scaling the consistency training \citep{song2023consistency, song2023improved} to text-to-image or higher resolution generation requires latent space training. Otherwise, with pretrained diffusion model, we could either finetune consistency training \citep{geng2024consistency} or distill from diffusion model \citep{song2023consistency, luo2023latent}. CM \citep{song2023consistency} is the first work proposing consistency distillation (CD) on pixel space. LCM \citep{luo2023latent} later applies consistency technique on latent space and can generate high-quality images within a few steps. However, LCM's generated images using 1-2 steps are still blurry \citep{luo2023latent}. Recent works, such as Hyper-SD \cite{ren2024hyper} and TCD \cite{zheng2024trajectory}, have introduced notable improvements to latent consistency distillation. TCD \cite{zheng2024trajectory} employed CTM \cite{kim2023consistency} instead of CD \cite{song2023consistency}, significantly enhancing the performance of the distilled student model. Building on this, Hyper-SD \cite{ren2024hyper} divided the Probability Flow ODE (PF-ODE) into multiple components inspired by Multistep Consistency Models (MCM) \cite{heek2024multistep}, and applied TCD \cite{zheng2024trajectory} to each segment. Subsequently, Hyper-SD \cite{ren2024hyper} merged these segments progressively into a final model, integrating human feedback learning and score distillation \cite{yin2024one} to optimize one-step generation performance.

\section{Preliminaries} \label{sec:bg}
Denote $\pdata(\rvx_0)$ as the data distribution, the forward diffusion process gradually adds Gaussian noise with monotonically increasing standard deviation $\sigma(t)$ for $t \in \{0,1,\dots,T\}$ such that $p_t(\rvx_t|\rvx_0) = \gN(\rvx_0, \sigma^2(t)\mI)$ and $\sigma(t)$ is handcrafted such that $\sigma(0) = \sigma_{\min}$ and $\sigma(T)=\sigma_{\max}$. By setting $\sigma(t) = t$, the probability flow ODE (PF-ODE) from \citep{Karras2022edm} is defined as:
\begin{equation}
    \frac{\rd\rvx_t}{\rd t} = -t\nabla_{\rvx_t} \log p_t(\rvx_t) = \frac{\left( \rvx_t - \vf(\rvx_t, t) \right)}{t},  \label{eq:pf_ode}
\end{equation}
where $\vf:(\rvx_t, t) \rightarrow \rvx_0$ is the denoising function which directly predicts clean data $\rvx_0$ from given perturbed data $\rvx_t$. 
\citep{song2023consistency} defines consistency model based on PF-ODE in \cref{eq:pf_ode}, which builds a bijective mapping $\vf$ between noisy distribution $p(\rvx_t)$ and data distribution $\pdata(\rvx_0)$. The bijective mapping $\vf:(\rvx_t, t) \rightarrow \rvx_0$ is termed the consistency function. A consistency model $\vf_\theta(\rvx_t, t)$ is trained to approximate this consistency function $\vf(\rvx_t, t)$. The previous works \citep{song2023consistency, song2023improved, Karras2022edm} impose the boundary condition by parameterizing the consistency model as:
\begin{equation}
    \vf_\theta(\rvx_t, t) = c_{skip}(t)\rvx_t + c_{out}(t)\mF_\theta(\rvx_t, t), \label{eq:cm_param}
\end{equation}
where $\mF_\theta(\rvx_t, t)$ is a neural network to train. Note that, since $\sigma(t) = t$, we hereafter use $t$ and $\sigma$ interchangeably. $c_{skip}(t)$ and $c_{out}(t)$ are time-dependent functions such that $c_{skip}(\sigma_{\min}) = 1$ and $c_{out}(\sigma_{\max}) = 0$.

To train or distill consistency model, \citep{song2023consistency, song2023improved, Karras2022edm} firstly discretize the PF-ODE using a sequence of noise levels $\sigma_{\min} = t_{\min} = t_1 < t_2 < \dots < t_{N} = t_{\max} = \sigma_{\max}$, where $t_i = \left( t_{\min}^{1/\rho} + \frac{i-1}{N-1}(t_{\max}^{1/\rho
} - t_{\min}^{1/\rho})\right)^\rho$ and $\rho = 7$. 

\textbf{Consistency Distillation} Given the pretrained diffusion model $\vs_\phi(\rvx_t, t) \approx \nabla_{\rvx_t} \log p_t(\rvx_t)$, the consistency model could be distilled from the pretrained diffusion model using the following CD loss:
\begin{equation}
    \gL_{\text{CD}}(\theta, \theta^-) = \E\left[ \lambda(t_i)d(\vf_\theta(\rvx_{t_{i+1}}, t_{i+1}), \vf_{\theta^{-}}(\Tilde{\rvx}_{t_i}, t_{i})) \right], \label{loss:cd}
\end{equation}
where $\rvx_{t_{i+1}} = \rvx_0 + t_{i+1} \rvz$ with the $\rvx_0 \sim \pdata(\rvx_0)$ and $\rvz \sim \gN(0, \mI)$ and $\rvx_{t_i} = \rvx_{t_{i+1}} - (t_{i}-t_{i+1})t_{i+1} \nabla_{\rvx_{t_{i+1}}} \log p_{t_{i+1}}(\rvx_{t_{i+1}}) = \rvx_{t_{i+1}} - (t_{i}-t_{i+1})t_{i+1}\vs_\phi(\rvx_{t_{i+1}}, t_{i+1})$. 

\textbf{Consistency Training}
The consistency model is trained by minimizing the following CT loss:
\begin{equation}
    \gL_{\text{CT}}(\theta, \theta^-) = \E\left[ \lambda(t_i)d(\vf_\theta(\rvx_{t_{i+1}}, t_{i+1}), \vf_{\theta^{-}}(\rvx_{t_i}, t_{i})) \right], \label{loss:ct}
\end{equation}
where $\rvx_{t_i} = \rvx_0 + t_{i} \rvz$ and $\rvx_{t_{i+1}} = \rvx_0 + t_{i+1} \rvz$ with the same $\rvx_0 \sim \pdata(\rvx_0)$ and $\rvz \sim \gN(0, \mI)$

In \cref{loss:cd} and \cref{loss:ct}, $\vf_\theta$ and $\vf_{\theta^-}$ are referred to as the online network and the target network, respectively. The target's parameter $\theta^-$ is obtained by applying the Exponential Moving Average (EMA) to the student's parameter $\theta$ during the training and distillation as follows:
\begin{equation}
    \theta^- \leftarrow \text{stopgrad}(\mu\theta^- + (1-\mu)\theta), \label{ema}
\end{equation}
with $0\leq\mu<1$ as the EMA decay rate,  weighting function $\lambda(t_i)$ for each timestep $t_i$, and $d(\cdot, \cdot)$ is a predefined metric function. 

In CM \citep{song2023consistency}, the consistency training still lags behind the consistency distillation and diffusion models. iCT \citep{song2023improved} later propose several improvements that significantly boost the training performance and efficiency. First, the EMA decay rate $\mu$ is set to $0$ for better training convergence. Second, the Fourier scaling factor of noise embedding and the dropout rate are carefully examined. Third, iCT introduces Pseudo-Huber losses to replace $L_2$ and LPIPS since LPIPS introduces the undesirable bias in generative modeling \citep{song2023improved}. Furthermore, the Pseudo-Huber is more robust to outliers since it imposes a smaller penalty for larger errors than the $L_2$ metric. Fourth, iCT proposes an exp curriculum for total discretization steps N, which doubles N after a predefined number of training iterations. Moreover, uniform weighting $\lambda(t_i) = 1$ is replaced by $\lambda(t_i)=1/(t_{i+1}-t_i)$. Finally, iCT adopts a discrete Lognormal distribution for timestep sampling as EDM \citep{Karras2022edm}. With all these improvements, CT is now better than CD and performs on par with the diffusion models in pixel space.

\section{Method}
\label{method}
In this paper, we first investigate the underlying reason behind the performance discrepancy between latent and pixel space using the same training framework in \cref{sec:analysis}. Based on the analysis, we find out the root of unsatisfied performance on latent space could be attributed to two factors: the impulsive outlier and the unstable temporal difference (TD) for computing consistency loss. To deal with impulsive outliers of TD on pixel space, \citep{song2023improved} proposes the Pseudo-Huber function as training loss. For the latent space, the impulsive outlier is even more severe, making Pseudo-Huber loss not enough to resist the outlier. Therefore,  \cref{sec:cauchy} introduces Cauchy loss, which is more effective with extreme outliers. In the next \cref{sec:diff_loss} and \cref{sec:ot}, we propose to use diffusion loss at early timesteps and OT matching for regularizing the overkill effect of consistency at the early step and training variance reduction, respectively. Section \ref{sec:c} designs an adaptive scheduler of scaling $c$ to control the robustness of the proposed loss function more carefully, leading to better performance. Finally, in \cref{sec:norm}, we investigate the normalization layers of architecture and introduce Non-scaling LayerNorm to both capture feature statistic better and reduce the sensitivity to outliers.

\subsection{Analysis of latent space} \label{sec:analysis}

We first reimplement the iCT model \citep{song2023improved} on the latent dataset CelebA-HQ $32 \times 32 \times 4$ and pixel dataset Cifar-10 $32 \times 32 \times 3$. Hereafter, we refer to the latent iCT model as iLCT. We find that iCT framework works well on pixel datasets as claim \citep{song2023improved}. However, it produces worse results on latent datasets as in \cref{fig:qualitative_ict} and \cref{tab:main_exp}. The iLCT gets a very high FID above 30 for both datasets, and the generative images are not usable in the real world. This observation raises concern about the sensitivity of CT algorithm with training data, and we should carefully examine the training dataset. In addition, we notice that the DQN and CM use the same TD loss, which update the current state using the future state. Furthermore, they also possess the training instability. This motivates to carefully examine the behavior of TD loss with different training data.

\begin{figure}[!t]
    \centering
    \includegraphics[width=0.85\linewidth]{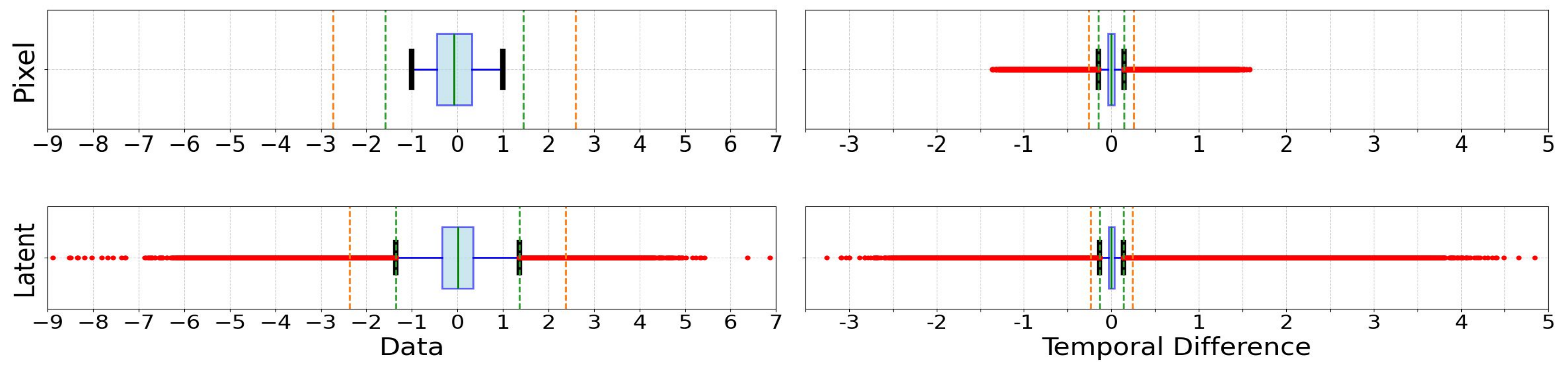}
    \caption{\textbf{Box and Whisker Plot:} Impulsive noise comparison between pixel and latent spaces. The right column shows the statistics of TD values at 21 discretization steps. Other discretization steps exhibit same behavior, where impulsive outliers are consistently present regardless of the total discretization steps. The blue boxes represent interquartile ranges of the data, while the green and orange dashed lines indicate inner and outer fences, respectively. Outliers are marked with red dots.}
    \label{fig:impulsive_noise}
\end{figure}

While the pixel data lies within the range $[-1, 1]$ after being normalized, the range of latent data varies depending on the encoder model, which is blackbox and unbound. After normalizing latent data using mean and variance, we observe that the latent data contains high-magnitude values. We call them the impulsive outliers since they account for small probability but are usually very large values. In the bottom left of \cref{fig:impulsive_noise}, the impulsive outlier of latent data is red, spanning from $-9$ to $7$, while the first and third quartiles are just around $-1.4$ and $1.4$, respectively. We evaluate how the iCT will be affected by data outliers by analyzing the temporal difference $\text{TD} = f_\theta(\rvx_{t_{i+1}}, t_{i+1})-f_{\theta^-}(\rvx_{t_i}, t_{i})$. In the top right of \cref{fig:impulsive_noise}, the impulsive outliers of pixel TD range from -1.5 to 1.7, which are not too far from the interquartile range compared to latent TD. The impulsive outliers of latent TD range is much wider from -3.2 to 5. iCT uses Pseudo-Huber loss instead of $L_2$ loss since the Huber is less sensitive to outliers, see \cref{fig:loss}. However, for latent data, the Huber's reduction in sensitivity to outliers is not enough. This indicates that even using Pseudo-Huber loss, the iLCT training on latent space could still be unstable and lead to worse performance, which matches our experiment results on iLCT. Based on the above analysis, we hypothesize that the TD value statistic highly depends on the training data statistic.

To mitigate the impact of impulsive outliers, we could use more stable target updates like Polyak or periodic in TD loss \cite{lee2019target}, but they lead to very slow convergence, as shown in \citep{song2023consistency}. Even though CM is initialized by a pretrained diffusion model, the Polyak update still takes a long time to converge. Therefore, using Polyak or periodic updates is computationally expensive, and we keep the standard target update as in \citep{song2023improved}. Another direction is using a special metric for latent like LPIPS on pixel space \citep{song2023consistency}. \citep{kang2024diffusion2gan} proposes the E-LatentLPIPS as a metric for distillation and performs well on distillation tasks. However, this requires training a network as a metric and using this metric during the training process will also increase the training budget. To avoid the overhead of the training, we seek a simple loss function like Pseudo-Huber but be more effective with outliers. We find that the Cauchy loss function \citep{black1996robust, barron2019general} could be a promising candidate in place of Pseudo-Huber for latent space.
\subsection{Cauchy Loss against Impulsive Outlier} \label{sec:cauchy}
\begin{figure}[!ht]
    \centering
    \begin{subfigure}[t]{0.40\textwidth}
        \centering
        \includegraphics[width=1.0\textwidth]{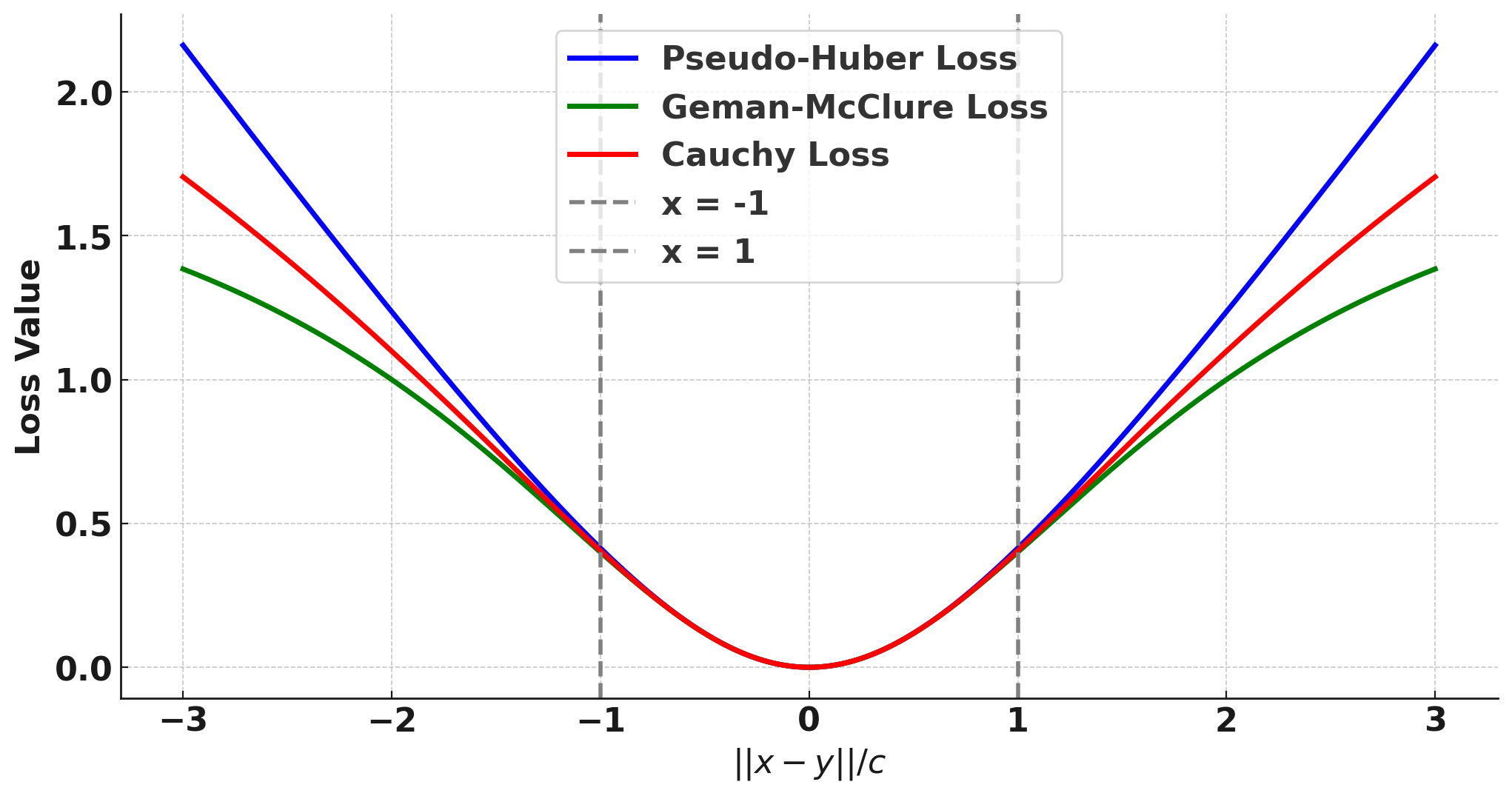}
        \caption{Robust Loss}
        \label{fig:loss_val}
    \end{subfigure}%
    ~ 
    \begin{subfigure}[t]{0.40 \textwidth}
        \centering
        \includegraphics[width=1.0\textwidth]{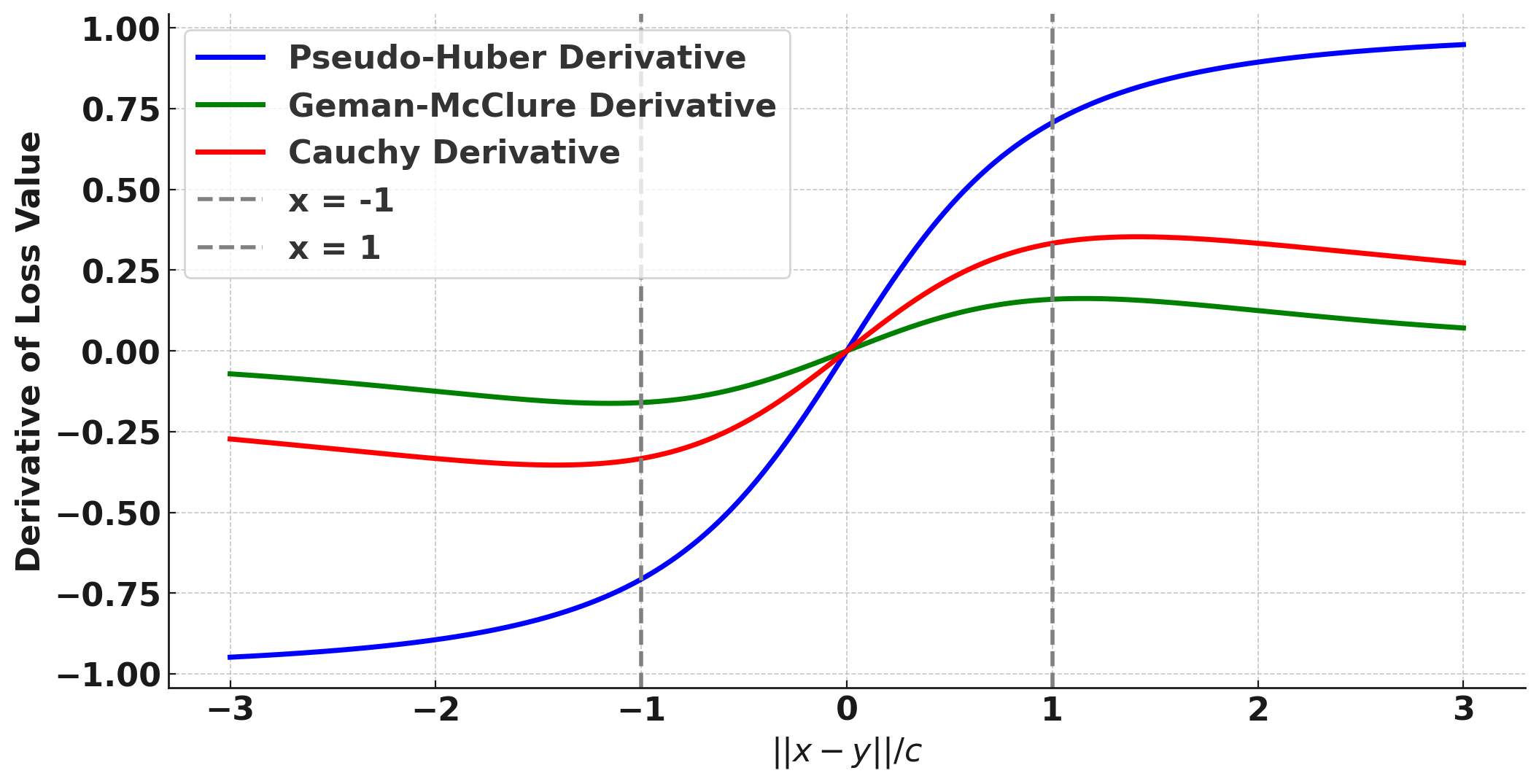}
        \caption{Derivative of Robust Loss}
        \label{fig:loss_derivative}
    \end{subfigure}
    \caption{Analysis of robust loss: Pseudo-Huber, Cauchy, and Geman-McClure}
    \label{fig:loss}
\end{figure}

In this section, we introduce the Cauchy loss \citep{black1996robust, barron2019general} function to deal with extreme impulsive outliers. The Cauchy loss function has the following form:
\begin{equation}
    d_{\text{Cauchy}}(\rvx, \rvy)=  \log \left(1+\frac{||\rvx-\rvy||_2^2}{2c^2}\right), \label{loss:cauchy}
\end{equation}
and we also consider two additional robust losses, which are Pseudo-Huber \citep{song2023improved, barron2019general} and Geman-McClure \citep{geman1986bayesian, barron2019general}
\begin{equation}
    d_{\text{Pseudo-Huber}}(\rvx, \rvy)= \sqrt{||\rvx-\rvy||_2^2 + c^2} - c, \label{loss:huber}
\end{equation}
\begin{equation}
    d_{\text{Geman-McClure}}(\rvx, \rvy)= \frac{2||\rvx-\rvy||_2^2}{||\rvx-\rvy||_2^2 + 4c^2}, \label{loss:gm}
\end{equation}
where $c$ is the scaling parameter to control how robust the loss is to the outlier. We analyze their robustness behavior against outliers. As shown in \cref{fig:loss_val}, the Pseudo-Huber loss linearly increases like $L_1$ loss for the large residuals $\rvx-\rvy$. In contrast, the Cauchy loss only grows logarithmically, and the Geman-McClure suppresses the loss value to $1$ for the outliers. 

The Pseudo-Huber loss works well if the residual value does not grow too high and, therefore, has a good performance on the pixel space. However, for the latent space, as shown in the bottom right of \cref{fig:impulsive_noise}, the TD suffers from extremely high values coming from the impulsive outlier in the latent dataset, the Cauchy loss could be more suitable since it significantly dampens the influence of extreme outliers. Otherwise, even Geman-McClure is very highly effective for removing outlier effects than two previous losses; it gives a gradient $0$ for high TD value and completely ignores the impulsive outliers as \cref{fig:loss_derivative}. This is unexpected behavior because even though we call the high-value latent impulsive outlier, they actually could encode important information from original data. Completely ignoring them could significantly hurt the performance of training model. Based on this analysis, we choose Cauchy loss as the default loss for latent CM for the rest of the paper. The loss ablation is provided in \cref{tab:ablate_robust}.

\subsection{Diffusion Loss at small timestep} \label{sec:diff_loss}
For small noise level $\sigma$, the ground truth of $f(\rvx_\sigma, \sigma)$ can be well approximated by $\rvx_0$, but this does not hold for large noise levels. Therefore, for low-level noise, the consistency objective seems to be overkill and harms the model's performance since instead of optimizing $f_\theta(\rvx_\sigma, \sigma)$ to approximated ground truth $\rvx_0$, the consistency objective optimizes through a proxy estimator $f_{\theta^-}(\rvx_{<\sigma}, <\sigma)$ leading to error accumulation over timestep. To regularize this overkill, we propose to apply an additional diffusion loss on small noise level as follows:

\begin{equation}
    L_{diff} = ||f_\theta(\rvx_{t_i}, t_i) - \rvx_0||^2_2 \quad \forall i \leq \text{int(N $\cdot$ r)}, \label{loss:diff}
\end{equation}

where N is the number of training discretization steps and $r\in[0;1]$ is the diffusion threshold, and we heuristicly choose $r=0.25$. We do not apply diffusion loss for large noise levels since $f(\rvx_\sigma, \sigma)$ will differ greatly from the target $\rvx_0$, leading to very high $L_2$ diffusion loss. This could harm the training consistency process, misleading to the wrong solution. We provide the ablation study in \cref{tab:diff_loss}. Furthermore, CTM \citep{kim2023consistency} also proposes to use diffusion loss, but they use them on both high and low-level noise, which is different from us. 

\subsection{OT matching reduces the variance} \label{sec:ot}
In this section, we adopt the OT matching technique from previous works \citep{pooladian2023multisample, lee2023minimizing}. \citep{pooladian2023multisample} proposes to use OT to match noise and data in the training batch, such as the moving $L_2$ cost is optimal. On the other hand, \citep{lee2023minimizing} introduces $\beta\text{VAE}$ for creating noise corresponding to data and train flow matching on the defined data-noise pairs. By reassigning noise-data pairs, these works significantly reduce the variance during the diffusion/flow matching training process, leading to a faster and more stable training process. According to \citep{zhang2023emergence}, the consistency training and diffusion models produce highly similar images given the same noise input. Therefore, the final output solution of the consistency and diffusion models should be close to each other. Since OT matching helps reduce the variance during training diffusion, it could be useful to reduce the variance of consistency training. In our implementation, we follow \citep{pooladian2023multisample, tong2023improving} using the POT library to map from noise to data in the training batch. The overhead caused by minibatch OT is relatively small, only around $0.93\%$ training time, but gains significant performance improvement as shown in \cref{tab:strategy}.

\subsection{Adaptive $c$ scheduler} \label{sec:c}


\begin{figure}[h!]
    \centering
    \includegraphics[width=0.8\linewidth]{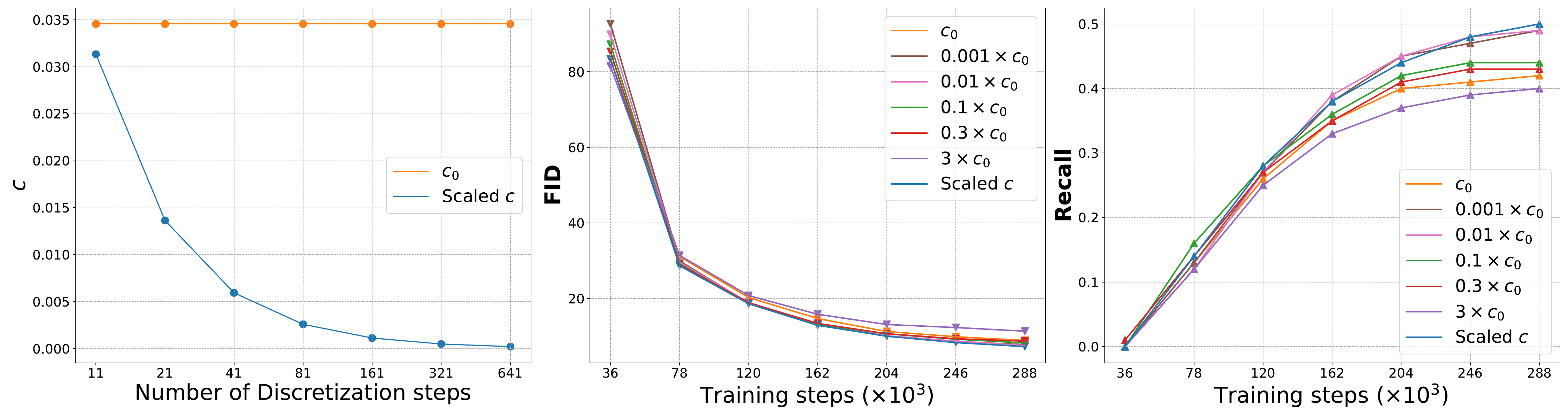}
    \caption{Model convergence plot on different $c$ schedule. (Left) Our proposed $c$ values. Performance on FID (Middle) and Recall (Right) of our proposed $c$ in comparison with different choices.}
    \label{fig:fid_vary_c}
\end{figure}

In this section, we examine the choice of scaling parameter $c$ in robust loss functions. The scaling parameter controls the robustness level, which is very important for model performance. The previous work \citep{song2023improved} proposes to use fixed constant $c_0 = 0.00054\sqrt{d}$, where $d$ is the dimension of data. We find that using this simple fixed $c$ is not yet optimal for the training consistency model. Especially in this paper, we follow the Exp curriculum specified by \cref{exp_cur} in \citep{song2023improved}, which doubles the total discretization step after a defined number of training iterations. 
\begin{equation}
    \text{NFE}(k)=\min \left(s_0 2^{\left\lfloor\frac{k}{K^{\prime}}\right\rfloor}, s_1\right)+1, \quad K^{\prime}=\left\lfloor\frac{K}{\log _2\left\lfloor s_1 / s_0\right\rfloor+1}\right\rfloor, \label{exp_cur}
\end{equation}
where $k$ is current training iteration, $K$ is total training iteration and $s_0 = 10, s_1=640$. During training, we notice that the variance of TD is significantly reduced as doubling total discretization steps using \cref{exp_cur}. Since the more discretization steps, the closer distance of $\rvx_{t_i}$ and $\rvx_{t_{i+1}}$, the TD value's range between them should be smaller. However, the impulsive outlier still exists regardless of the number of discretization steps. Intuitively, we propose a heuristic adaptive $c$ scheduler where the $c$ is scaled down proportional to the reduction rate of TD variance as the number of discretization steps increases. We plot our $c$ scheduler versus discretization steps in \cref{fig:fid_vary_c} and we fit the $c$ scheduler to get the scheduler equation as following:

\begin{equation}
    c = \exp(-1.18 * \log(\text{NFE}(k) - 1) - 0.72) \label{eq:c_scheduler}
\end{equation}

\subsection{Non-scaling Layernorm} \label{sec:norm}
As mentioned in \cref{sec:analysis}, the statistic of training data could play an important role in the success of consistency training. Furthermore, in architecture design, the normalization layer specifically handles the statistics of input, output, and hidden features. In this section, we investigate the normalization layer choice for consistency training, which is sensitive to training data statistics. 

Currently, both \citep{song2023improved, song2023consistency} use the UNet architecture from \citep{dhariwal2021diffusion}. In UNet \citep{dhariwal2021diffusion}, GroupNorm is used in every layer by default. The GroupNorm only captures the statistics over groups of local channels, while the LayerNorm further captures the statistics' overall features. Therefore, LayerNorm is better at capturing fine-grained statistics over the entire feature. We further carry out the experiments for other types of normalization, such as LayerNorm, InstanceNorm, RMSNorm in \cref{tab:norm_layer} and observe that the GroupNorm and InstanceNorm perform relatively well compared to others, especially LayerNorm. This could be due to that they are less sensitive to the outliers since they only capture the statistic over groups of channels. Therefore, the impulsive features only affect the normalization of a group containing them. For the LayerNorm, the impulsive features could negatively impact the overall features's normalization. We further look into the LayerNorm implementation and suspect that the scaling term could significantly amplify the outliers across features by serving as a shared parameter. This observation is also mentioned in \citep{wei2022outlier} for LLM quantization. In implementation, we set the \textbf{scaling term of LayerNorm to $1$} and \textbf{disabled the gradient update} for it \eqref{operation:layernorm}. We refer to it as Non-scaling LayerNorm (NsLN) as \citep{wei2022outlier}.

\begin{equation}
    \text{LN}_{\gamma, \beta}(\rvx) = \frac{\rvx - u(\rvx)}{\sqrt{\sigma^{2}(\rvx) + \epsilon}} \cdot \gamma + \beta, \quad
    \text{NsLN}_{\beta}(\rvx) = \frac{\rvx - u(\rvx)}{\sqrt{\sigma^{2}(\rvx) + \epsilon}} + \beta, \label{operation:layernorm}
\end{equation}

where $u(\rvx)$ and $\sigma^{2}(\rvx)$ are mean and variance of $\rvx$.


\section{Experiment} \label{exp}

\subsection{Performance of our training technique} \label{exp:main}
\begin{table}[t]
    \centering
    \begin{tabular}{cc}
        \begin{minipage}[c]{0.58\textwidth}
            \centering
            \begin{subtable}[t]{\textwidth}
                \resizebox{\textwidth}{!}{%
                \begin{tabular}{l c c c c c}
                    \toprule
                    Model & NFE$\downarrow$ & FID$\downarrow$ & Recall$\uparrow$ & Epochs & Total Bs\\
                    \midrule 
                    \multicolumn{5}{c}{\textbf{Pixel Diffusion Model}}\\
                    \midrule
                    WaveDiff \citep{phung2023wavediff} & 2 & 5.94 & 0.37 & 500 & 64\\
                    Score SDE \citep{song2020score} & 4000 & 7.23 & - & ~6.2K & - \\
                    DDGAN \citep{xiao2021tackling} & 2 & 7.64 & 0.36 & 800 & 32 \\
                    RDUOT \citep{dao2024high} & 2 & 5.60 & 0.38 & 600 & 24 \\
                    RDM \citep{teng2023relay} & 270 & 3.15 & 0.55 & 4K & - \\
                    UNCSN++ \citep{kim2021soft} & 2000 & 7.16 & - & - & -\\
                    \midrule 
                    \multicolumn{5}{c}{\textbf{Latent Diffusion Model}}\\
                    \midrule
                    LFM-8 \citep{dao2023flow} & 85 & 5.82 & 0.41 & 500 & 112\\ 
                    LDM-4 \citep{rombach2021highresolution} & 200 & 5.11 & 0.49 & 600 &48 \\
                    LSGM \citep{vahdat2021score} & 23 & 7.22 & - & 1K &-\\
                    DDMI \citep{park2024ddmi} & 1000 & 7.25 & - & - &-\\
                    
                    DIMSUM \citep{phung2024dimsum} & 73 & 3.76  & 0.56 & 395 &32\\
                    $\text{LDM-8}^\dagger$ & 250 & {8.85}  & - & 1.4K &128\\
                    
                    \midrule
                    \multicolumn{5}{c}{\textbf{Latent Consistency Model}}\\
                    \midrule
                    iLCT \citep{song2023improved} & 1 & 37.15 & 0.12 & 1.4K &128\\
                    iLCT \citep{song2023improved} & 2 & 16.84 & 0.24 & 1.4K &128\\
                    Ours  & 1 & 7.27 & 0.50 & 1.4K &128\\
                    Ours  & 2 & 6.93 & 0.52 & 1.4K &128\\
                    \bottomrule
                \end{tabular}%
                }
            \caption{CelebA-HQ}
            \label{tab:celeb}
            \end{subtable}
        \end{minipage}
        \hfill
        \begin{minipage}[c]{0.42\textwidth}
            \centering
            \begin{subtable}[t]{\textwidth}
                \resizebox{\textwidth}{!}{%
                \begin{tabular}{l c c c c c}
                    \toprule
                    Model & NFE$\downarrow$ & FID$\downarrow$ & Recall$\uparrow$ & Epochs & Total Bs\\
                    \midrule 
                    \multicolumn{5}{c}{\textbf{Pixel Diffusion Model}}\\
                    \midrule
                    WaveDiff \citep{phung2023wavediff} & 2 & 5.94 & 0.37 & 500 & 64\\
                    Score SDE \citep{song2020score} & 4000 & 7.23 & - &6.2K & -\\
                    DDGAN \citep{xiao2021tackling} & 2 & 5.25 & 0.36 & 500 & 32\\
                    \midrule 
                    \multicolumn{5}{c}{\textbf{Latent Diffusion Model}}\\
                    \midrule
                    LFM-8 \citep{dao2023flow} & 90 & 7.70 & 0.39 & 90 &112\\
                    LDM-8 \citep{rombach2021highresolution} & 400 & 4.02 & 0.52 & 400 &96\\
                    $\text{LDM-8}^\dagger$ & 250 & {10.81} & - & 1.8K &256\\
                    \midrule
                    \multicolumn{5}{c}{\textbf{Latent Consistency Model}}\\
                    \midrule
                    iLCT \citep{song2023improved} & 1 &52.45  &0.11  & 1.8K &256\\
                    iLCT \citep{song2023improved} & 2 &24.67  &0.17  & 1.8K &256\\
                    Ours  & 1 &8.87  &0.47  & 1.8K &256\\
                    Ours  & 2 &7.71  &0.48  & 1.8K &256\\
                    \bottomrule
                \end{tabular}%
                }
            \caption{LSUN Church}
            \label{tab:lsun}
            \end{subtable}
            \hfill
            \begin{subtable}[t]{\textwidth}
                \resizebox{\textwidth}{!}{%
                \begin{tabular}{l c c c c c}
                    \toprule
                    Model & NFE$\downarrow$ & FID$\downarrow$ & Recall$\uparrow$ & Epochs &Total Bs \\
                    \midrule 
                    \multicolumn{5}{c}{\textbf{Latent Diffusion Model}}\\
                    \midrule
                    LFM-8 \citep{dao2023flow} & 84 & 8.07 & 0.40 & 700 &128\\
                    LDM-4 \citep{rombach2021highresolution} & 200 & 4.98 & 0.50 & 400 &42\\
                    $\text{LDM-8}^\dagger$ & 250 &{10.23} & - & 1.4K &128\\
                    \midrule
                    \multicolumn{5}{c}{\textbf{Latent Consistency Model}}\\
                    \midrule
                    iLCT \citep{song2023improved} & 1 & 48.82  & 0.15 & 1.4K &128 \\
                    iLCT \citep{song2023improved} & 2 & 21.15 & 0.19 & 1.4K &128\\
                    Ours  & 1 & 8.72  &0.42 & 1.4K &128\\
                    Ours  & 2 & 8.29  &0.43  & 1.4K &128\\
                    \bottomrule
                \end{tabular}%
                }
            \caption{FFHQ}
            \label{tab:ffhq}
            \end{subtable}
        \end{minipage}
    \end{tabular}
    \caption{Our performance on CelebA-HQ, LSUN Church, FFHQ datasets at resolution $256 \times 256$. ($\dagger$) means training on our machine with the same diffusion forward and equivalent architecture.}
    \label{tab:main_exp}
\end{table}

\minisection{Experiment Setting:}
We measure the performance of our proposed technique on three datasets: CelebA-HQ \citep{celeba}, FFHQ \citep{karras2019style}, and LSUN Church \citep{lsun}, at the same resolution of $256 \times 256$. Following LDM \citep{rombach2021highresolution}, we use pretrained VAE KL-8 \footnote{https://huggingface.co/stabilityai/sd-vae-ft-ema} to obtain latent data with the dimensionality of $32 \times 32 \times 4$. We adopt the OpenAI UNet architecture \citep{dhariwal2021diffusion} as the default architecture throughout the paper. Furthermore, we use the variance exploding (VE) forward process for all the consistency and diffusion experiments following \citep{song2023consistency, song2023improved}. The baseline iCT is self-implemented based on official implementation CM \citep{song2023consistency} and iCT \citep{song2023improved}. We refer to this baseline as iLCT. Furthermore, we also train the latent diffusion model for each dataset using the same VE forward noise process for fair comparisons with our technique. This LDM model is referred to as $\text{LDM-8}^{\dagger}$ in \cref{tab:main_exp}. All three frameworks, including ours, iLCT, and $\text{LDM-8}^{\dagger}$, use the same architecture.

\minisection{Evaluation:} During the evaluation, we first generate 50K latent samples and then pass them through VAE's decoder to obtain the pixel images. We use two well-known metrics, Fréchet Inception Distance (FID) \citep{fid} and Recall \citep{kynkaanniemi2019improved}, for measuring the performance of the model given the training data and 50K generated images. 

\begin{figure}[ht]
\centering
    \begin{subfigure}[b]{0.3\textwidth}
    \centering
    \includegraphics[width=\textwidth]{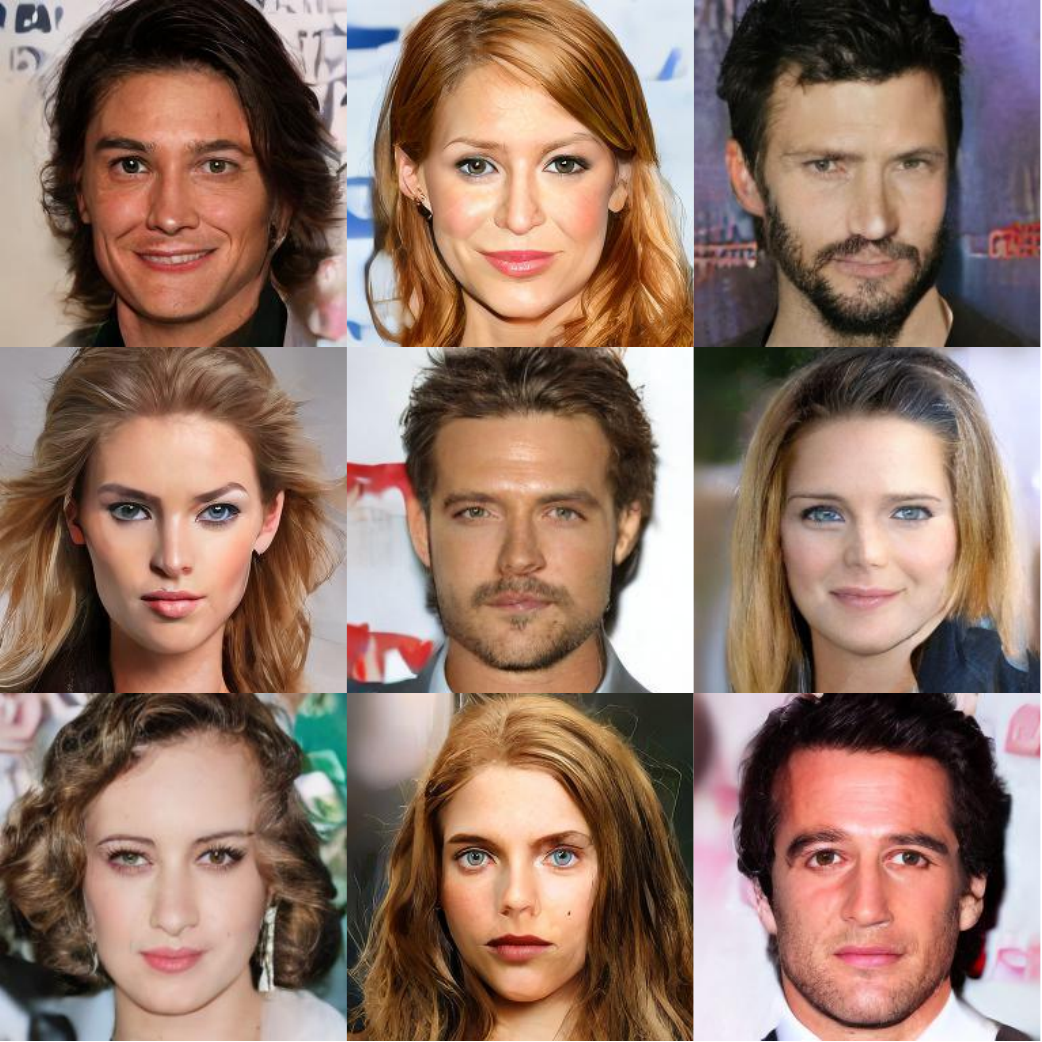}
    \caption{CelebA-HQ}
    \label{fig:qualitative_celeba}
    \end{subfigure}
    \hfill
    \begin{subfigure}[b]{0.3\textwidth}
    \centering
    \includegraphics[width=\textwidth]{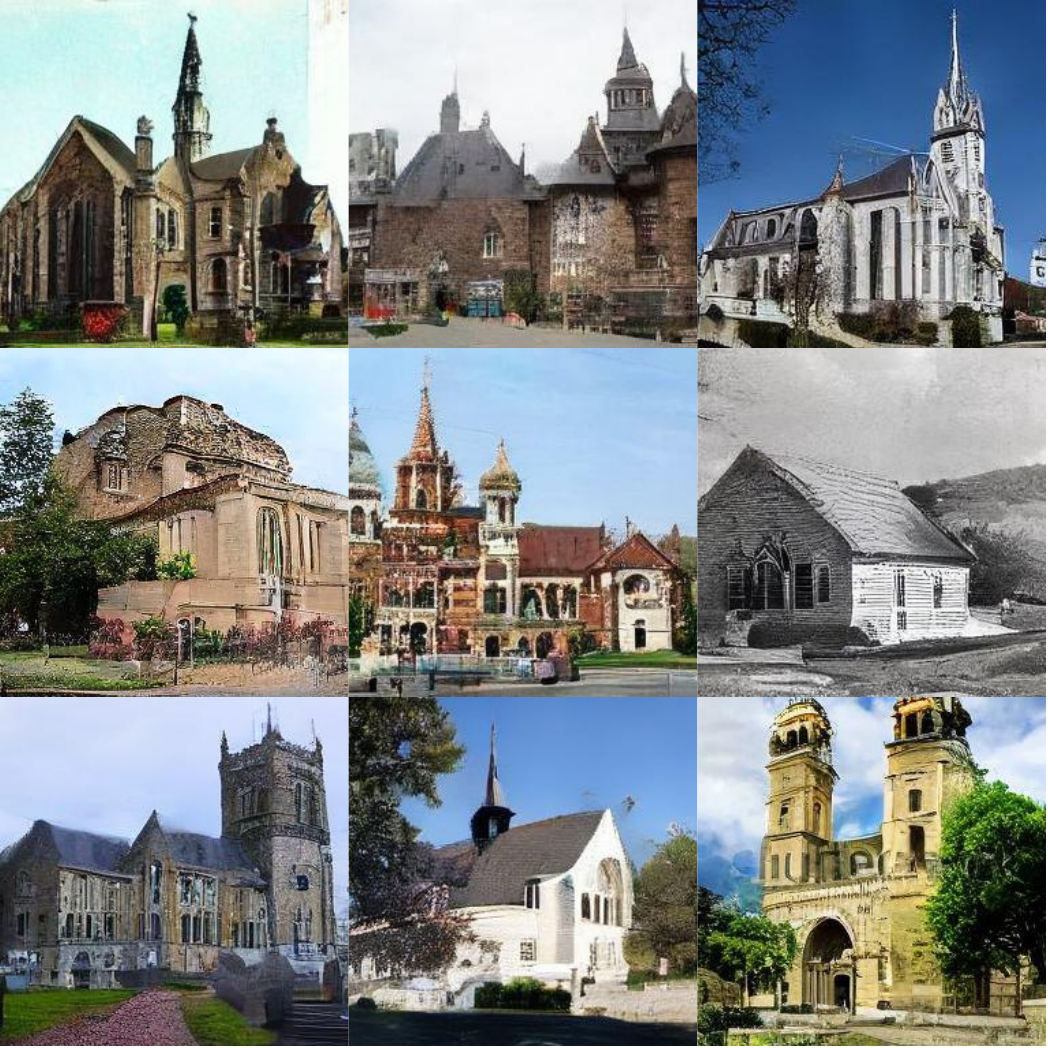}
    \caption{LSUN Church}
    \label{fig:qualitative_lsun_church}
    \end{subfigure}
    \hfill
    \begin{subfigure}[b]{0.3\textwidth}
    \centering
    \includegraphics[width=\textwidth]{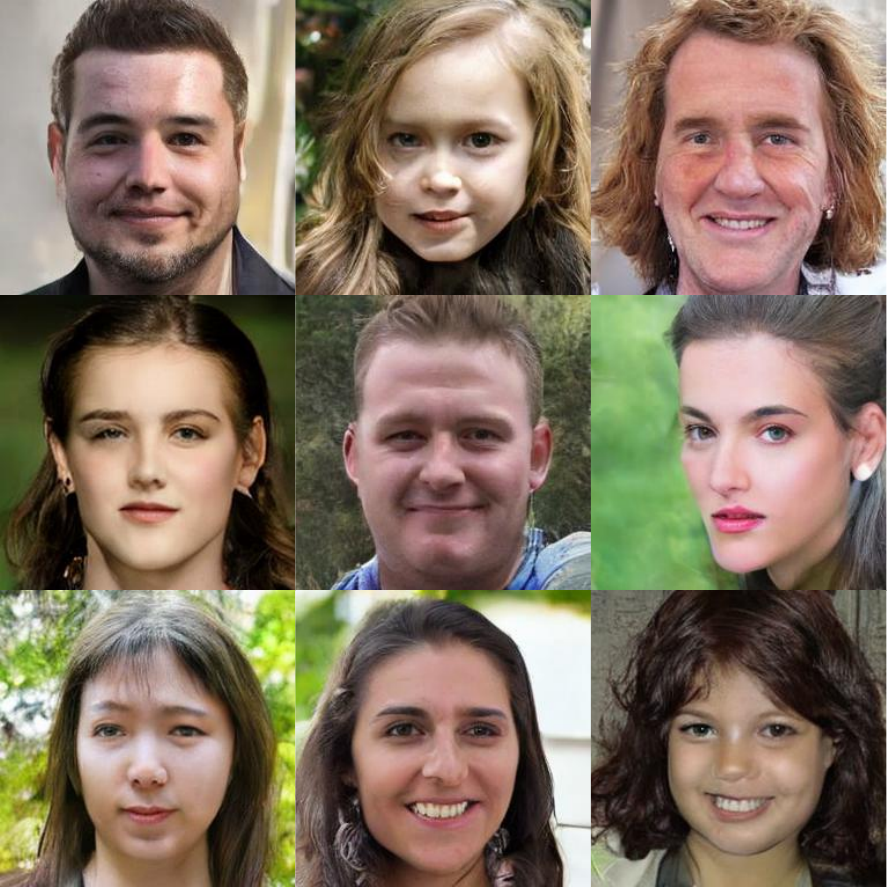}
    \caption{FFHQ}
    \label{fig:qualitative_ffhq}
    \end{subfigure}
    \caption{Our qualitative results using 1-NFE at resolution $256 \times 256$}
    \label{fig:qualitative_1nfe}
\end{figure}

\begin{figure}[ht]
\centering
    \begin{subfigure}[b]{0.3\textwidth}
    \centering
    \includegraphics[width=\textwidth]{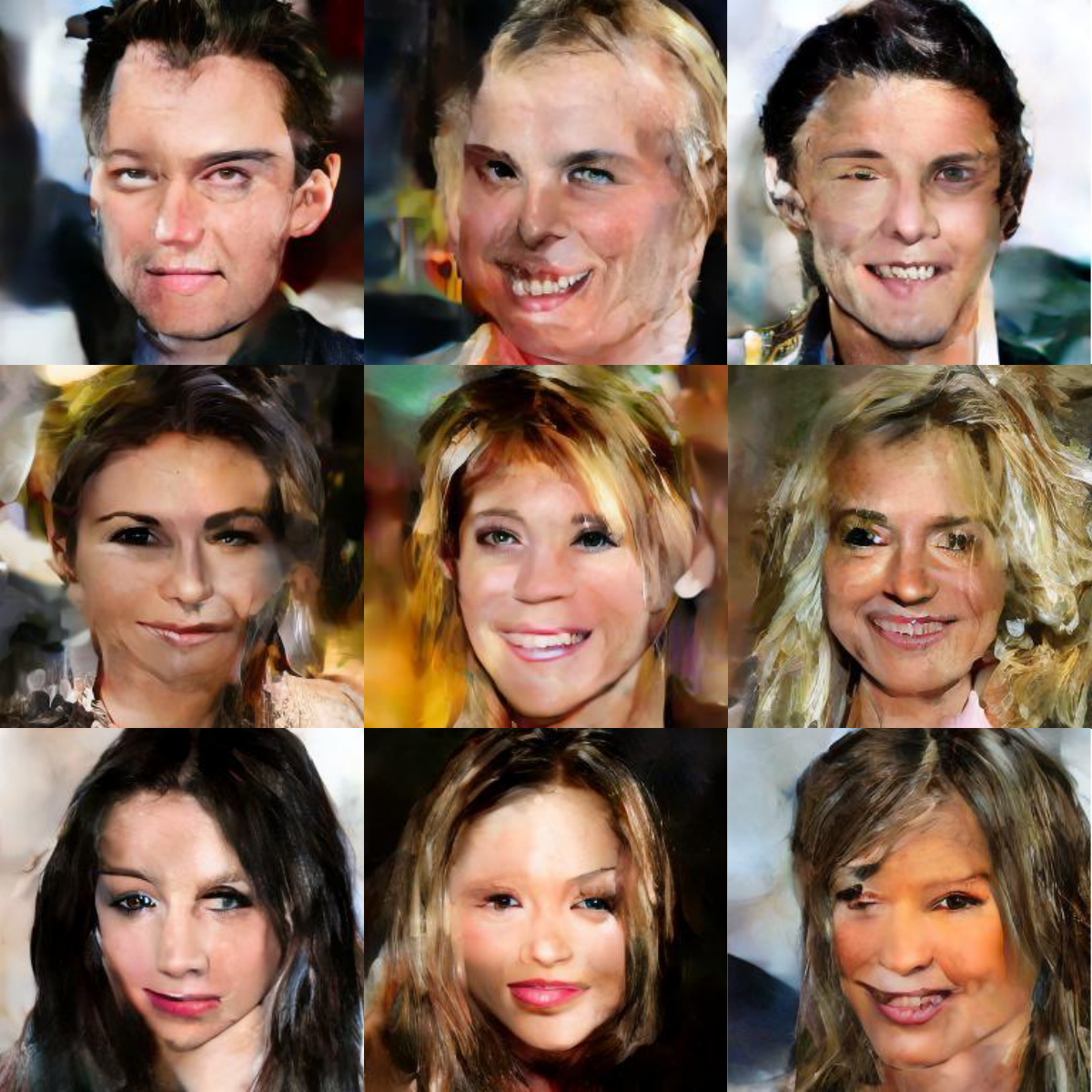}
    \caption{CelebA-HQ}
    \label{fig:qualitative_ict_celeba}
    \end{subfigure}
    \hfill
    \begin{subfigure}[b]{0.3\textwidth}
    \centering
    \includegraphics[width=\textwidth]{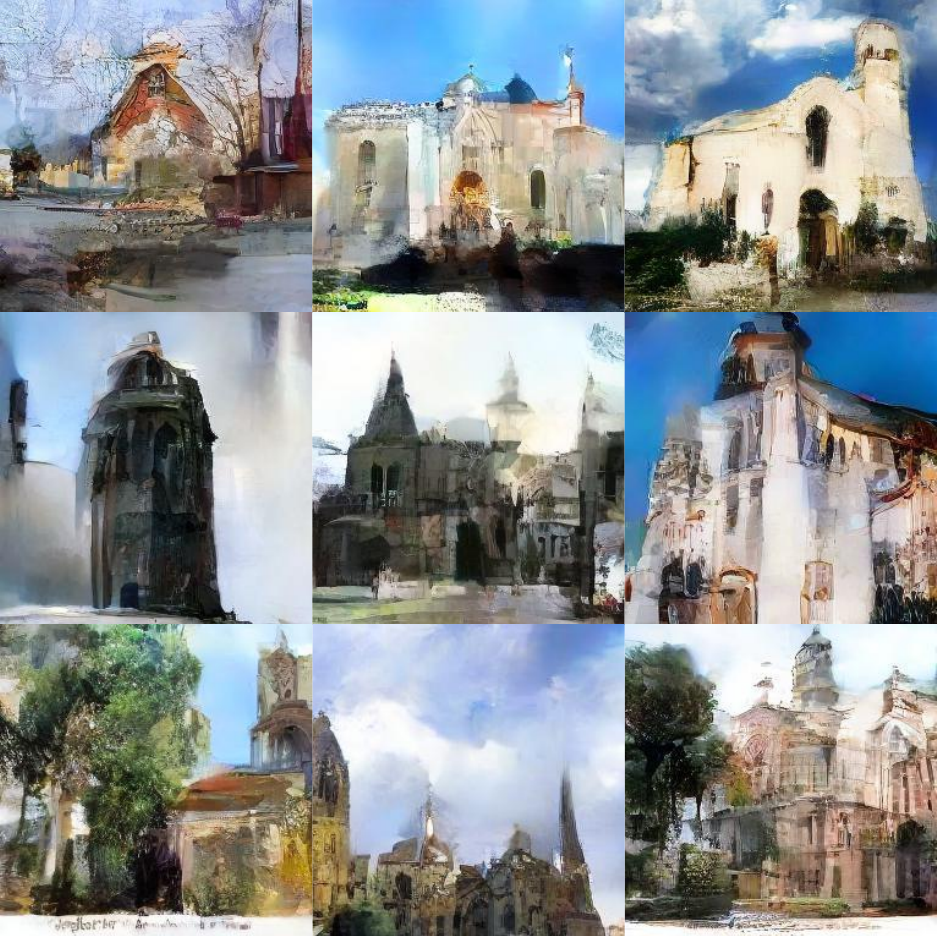}
    \caption{LSUN Church}
    \label{fig:qualitative_ict_lsun_church}
    \end{subfigure}
    \hfill
    \begin{subfigure}[b]{0.3\textwidth}
    \centering
    \includegraphics[width=\textwidth]{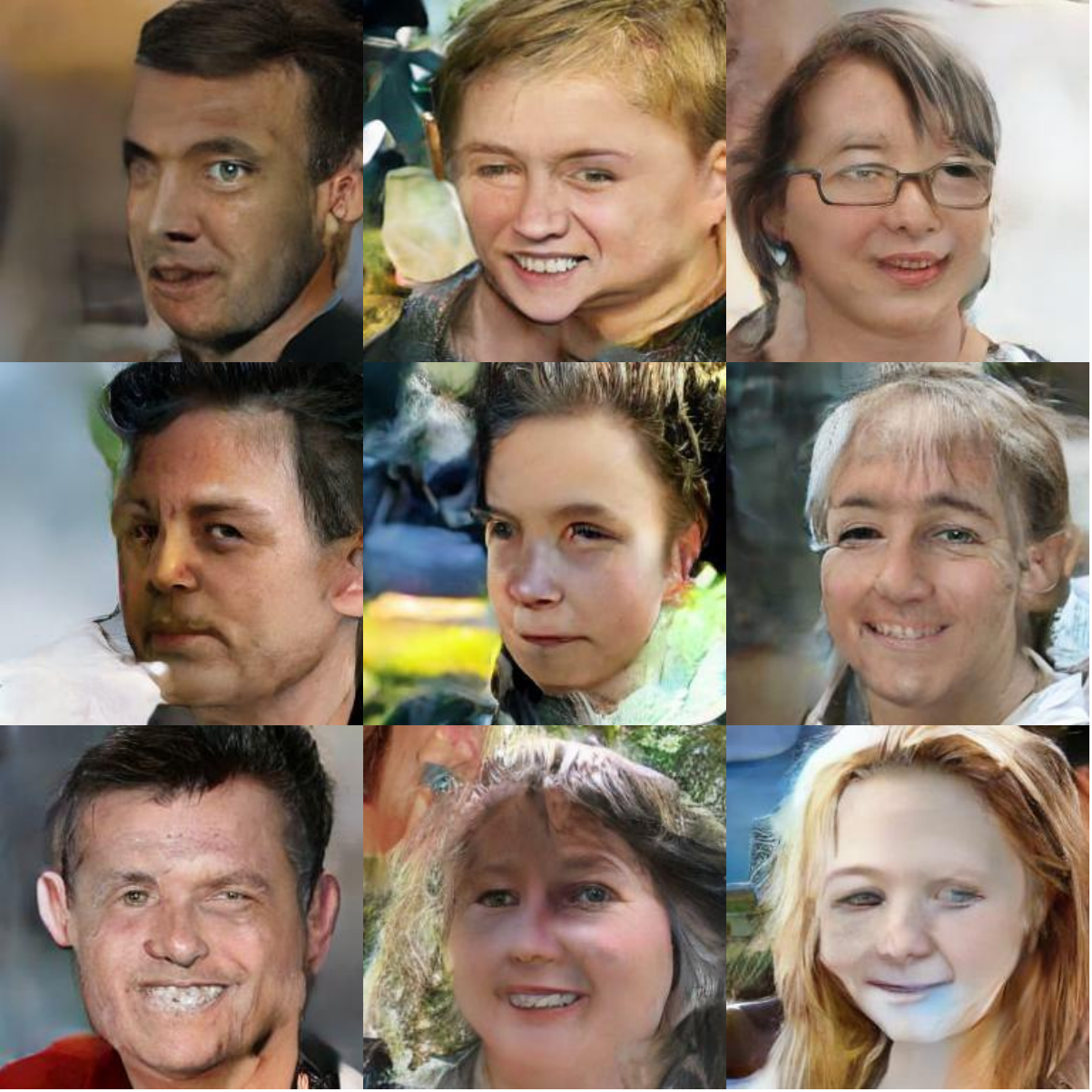}
    \caption{FFHQ}
    \label{fig:qualitative_ict_ffhq}
    \end{subfigure}
    \caption{iLCT qualitative results using 1-NFE at resolution  $256 \times 256$}
    \label{fig:qualitative_ict}
\end{figure}

\minisection{Model Performance:} We report the performance of our model across all three datasets in \cref{tab:main_exp}, primarily to compare it with the baseline iLCT \citep{song2023improved} and LDM \citep{rombach2021highresolution}. For both 1 and 2 NFE sampling, we observe that the FIDs of iLCT for all datasets are notably high (over 30 for 1-NFE sampling and over 16 for 2-NFE sampling), consistent with the qualitative results shown in \cref{fig:qualitative_ict}, where the generated image is unrealistic and contain many artifacts. This poor performance of iLCT in latent space is expected, as the Pseudo-Huber training losses are insufficient in mitigating extreme impulsive outliers, as discussed in \cref{sec:analysis} and \cref{sec:cauchy}. In contrast, our proposed framework demonstrates significantly better FID and Recall than iLCT. Specifically, we achieve 1-NFE sampling FIDs of 7.27, 8.87, and 8.29 for CelebA-HQ, LSUN Church, and FFHQ, respectively. For 2-NFE sampling, our FID scores improve across all three datasets. Notably, our 1-NFE sampling outperforms $\text{LDM-8}^{\dagger}$, using the same noise scheduler and architecture. However, our models still exhibit higher FIDs compared to LDM \citep{rombach2021highresolution} and LFM \citep{dao2023flow}. In contrast, we only need 1 or 2 timestep sampling, whereas they require multiple timesteps for high-fidelity generation.
 It's important to note that we employ the VE forward process, whereas these other methods use VP and flow-matching forward processes. Furthermore, the qualitative results of our framework, as shown in \cref{fig:qualitative_1nfe}, highlight our ability to generate high-quality images.

\subsection{Ablation of proposed framework} \label{exp:ablation}

We ablate our proposed techniques on the CelebA-HQ $256\times256$ dataset, with all FID and Recall metrics measured using 1-NFE sampling. All models are trained for 1,400 epochs with the same hyperparameters. As shown in \cref{tab:strategy}, replacing Pseudo-Huber losses with Cauchy losses makes our model's training less sensitive to impulsive outliers, resulting in a significant FID reduction from $37.15$ to $13.02$. This demonstrates the effectiveness of Cauchy losses in handling extremely high-value outliers, as discussed in \cref{sec:cauchy}. Additionally, applying diffusion loss at small timesteps further reduces FID by approximately 4 points to $9.11$, as this loss term stabilizes the training process at small timesteps, as described in \cref{sec:diff_loss}. Introducing OT coupling during minibatch training reduces training variance, improving the FID to $8.89$. Notably, by replacing the fixed scaling term $c=c_0$, \citep{song2023improved} with an adaptive scaling schedule, our model achieves an additional FID reduction of more than 1 point, reaching $7.76$, highlighting the importance of the scaling term $c$ in robustness control. Finally, we propose using NsLN, which removes the scaling term from LayerNorm to handle outliers more effectively. NsLN captures feature statistics while mitigating the negative impact of outliers, resulting in our best FID of $7.27$.

\minisection{Robustness Loss} \label{exp:ablation:robust_loss}
To analyze the impact of different robust loss functions, we conduct an ablation study using our best settings but replace the Cauchy loss with alternatives such as L2, E-LatentLPIPS \cite{kang2024diffusion2gan}, the Huber and the Geman-McClure loss. The results, shown in \cref{tab:ablate_robust}, indicate that both Huber and Geman-McClure underperform compared to the Cauchy loss when applied in the latent space. This is because the Huber loss remains too sensitive to extremely impulsive outliers, while the Geman-McClure loss tends to ignore such outliers entirely, leading to a loss of important information. This behavior is also discussed in \cref{sec:cauchy}.

\begin{table}[h!]
    \centering
    \begin{tabular}{cc}
        \begin{minipage}[c]{0.40\textwidth}
            \centering
            
            \begin{subtable}[t]{\textwidth}
                \centering
                \begin{tabular}{lcc}
                    \toprule
                    Framework                      & FID $\downarrow$   & Recall $\uparrow$   \\
                    \midrule
                    iLCT                           & 37.15              & 0.12                \\
                    \midrule
                    Cauchy                         & 13.02              & 0.36                \\
                    + Diff                         & 9.11               & 0.41                \\
                    + OT                           & 8.89               & 0.42                \\
                    + Scaled $c$                   & 7.76               & 0.47                \\
                    + NsLN       & \textbf{7.27}               &\textbf{0.50}                \\
                    \bottomrule
                \end{tabular}
                \caption{Components of proposed framework}
                \label{tab:strategy}
            \end{subtable}
            \hfill
            \begin{subtable}[t]{\textwidth}
                \centering
                \begin{tabular}{lcc}
                    \toprule
                    $r$           & FID $\downarrow$   & Recall $\uparrow$   \\
                    \midrule
                    1.0                    & 7.47               & 0.49                \\
                    0.6                      & 7.33               & 0.49                \\
                    0.25   & \textbf{7.27}               & \textbf{0.50}                \\
                    \bottomrule
                \end{tabular}
                \caption{Threshold using Diffusion loss}
                \label{tab:diff_loss}
            \end{subtable}
            
        \end{minipage}
        \hfill
        \begin{minipage}[c]{0.40\textwidth}
            \centering
            \begin{subtable}[t]{\textwidth}
                \centering
                \begin{tabular}{lcc}
                    \toprule
                    Loss                        & FID $\downarrow$   & Recall $\uparrow$   \\
                    \midrule
                     L2                          & 50.40              & 0.04                \\
                    E-LatentLPIPS               & 11.49              & 0.47                \\
                    \midrule
                    Huber                       & 9.97               & 0.44                \\
                    Geman McClure               & 11.28              & 0.44                \\
                    Cauchy   & \textbf{7.27}               & \textbf{0.50}                \\
                    \bottomrule
                \end{tabular}
                \caption{Robust losses.}
                \label{tab:ablate_robust}
            \end{subtable}
            \hfill
            \begin{subtable}[t]{\textwidth}
                \centering
                \begin{tabular}{lcc}
                    \toprule
                    Norm layer                             & FID $\downarrow$   & Recall $\uparrow$   \\
                    \midrule
                    $\text{GN}$                           & 7.76               & 0.47                \\
                    \midrule
                    IN                                      & 8.47               & 0.43                \\
                    LN                                      & 9.05               & 0.46                \\
                    RMS                                     & 8.96               & 0.46                \\
                    NsLN                 &\textbf{7.27}               &\textbf{0.50}                \\
                    \bottomrule
                \end{tabular}
                \caption{Norm Layer}
                \label{tab:norm_layer}
            \end{subtable}
        \end{minipage}
    \end{tabular}
    \caption{Ablation Studies on CelebA-HQ $256\times256$ dataset at epoch 1400}
    \label{tab:ablation}
\end{table}

\minisection{Diffusion Threshold} \label{exp:ablation:diff_loss}
In this section, we explore the impact of varying the threshold for applying the diffusion loss function in combination with the consistency loss. We observe that using the diffusion loss at every timestep improves consistency training; however, it underperforms compared to applying the diffusion loss selectively at smaller timesteps such as $r=0.25$ as shown in \cref{tab:diff_loss}. This suggests that applying diffusion losses primarily at small noise levels improves performance as discussed \cref{sec:diff_loss}. At larger timesteps, the diffusion loss may conflict with the consistency loss, potentially guiding the model toward incorrect solutions, thereby reducing overall performance.

\minisection{Scaling term $c$ scheduler} \label{exp:ablation:vary_c}
In this section, we compare the performance of our adaptive scaling $c$ scheduler with the fixed scaling $c$ scheduler proposed in \citep{song2023improved}. Our model demonstrates better convergence with the proposed adaptive $c$ scheduler. The rationale behind this improvement lies in the fact that, as the discretization steps increases using the exponential curriculum, the value of the TD scales down. Despite the reduced TD value, impulsive outliers still persist. A fixed large scaling $c$ is not effective in handling these outliers. To address this, we scale $c$ down as discretization steps increases, which leads to better performance, as shown in \cref{fig:fid_vary_c}.

\minisection{Normalizing Layer} \label{exp:ablation:norm_layer}
We denote GN, IN, LN, RMS, and NsLN as GroupNorm, InstanceNorm, LayerNorm, RMSNorm, and Non-scaling LayerNorm, respectively. The baseline UNet architecture from \citep{dhariwal2021diffusion} uses GroupNorm by default. We replace the normalization layers in the baseline with each of these types and train the model on CelebA-HQ using the best settings. The results are reported in \cref{tab:norm_layer}. GN and IN only capture local statistics, making them more robust to outliers, as outliers in one region do not affect others. In contrast, LN captures statistics from all features, making it more vulnerable to outliers because an outlier affects all features through a shared scaling term. By removing the scaling term in LN, we obtain NsLN, which is both effective in capturing feature statistics and resistant to outliers. As shown in \cref{tab:norm_layer}, NsLN outperforms the second-best GN by 0.5 FID and significantly outperforms LN.
\section{Conclusion}
CT is highly sensitive to the statistical properties of the training data. In particular, when the data contains impulsive noise, such as latent data, CT becomes unstable, leading to poor performance. In this work, we propose using the Cauchy loss, which is more robust to outliers, along with several improved training strategies to enhance model performance. As a result, we can generate high-fidelity images from latent CT, effectively bridging the gap between latent diffusion models and consistency models. Future work could explore further improvements to the architecture, specifically by investigating normalization methods that reduce the impact of outliers. For example, removing the scaling term from group normalization or instance normalization may help mitigate outlier effects. Another promising future direction is the integration of this technique with Consistency Trajectory Models (CTM) \cite{kim2023consistency}, as CTM has demonstrated improved performance compared to traditional Consistency Models (CM) \cite{song2023consistency}.
\section*{Acknowledgements}
Research funded by research grants to Prof. Dimitris Metaxas from NSF: 2310966, 2235405, 2212301, 2003874, 1951890, AFOSR 23RT0630, and NIH 2R01HL127661.

\bibliography{iclr2025_conference}

\begin{thebibliography}{60}
\providecommand{\natexlab}[1]{#1}
\providecommand{\url}[1]{\texttt{#1}}
\expandafter\ifx\csname urlstyle\endcsname\relax
  \providecommand{\doi}[1]{doi: #1}\else
  \providecommand{\doi}{doi: \begingroup \urlstyle{rm}\Url}\fi

\bibitem[Barron(2019)]{barron2019general}
Jonathan~T Barron.
\newblock A general and adaptive robust loss function.
\newblock In \emph{Proceedings of the IEEE/CVF conference on computer vision and pattern recognition}, pp.\  4331--4339, 2019.

\bibitem[Berthelot et~al.(2023)Berthelot, Autef, Lin, Yap, Zhai, Hu, Zheng, Talbott, and Gu]{berthelot2023tract}
David Berthelot, Arnaud Autef, Jierui Lin, Dian~Ang Yap, Shuangfei Zhai, Siyuan Hu, Daniel Zheng, Walter Talbott, and Eric Gu.
\newblock Tract: Denoising diffusion models with transitive closure time-distillation.
\newblock \emph{arXiv preprint arXiv:2303.04248}, 2023.

\bibitem[Black \& Anandan(1996)Black and Anandan]{black1996robust}
Michael~J Black and Paul Anandan.
\newblock The robust estimation of multiple motions: Parametric and piecewise-smooth flow fields.
\newblock \emph{Computer vision and image understanding}, 63\penalty0 (1):\penalty0 75--104, 1996.

\bibitem[Brooks et~al.(2023)Brooks, Holynski, and Efros]{brooks2022instructpix2pix}
Tim Brooks, Aleksander Holynski, and Alexei~A. Efros.
\newblock Instructpix2pix: Learning to follow image editing instructions.
\newblock In \emph{CVPR}, 2023.

\bibitem[Dao et~al.(2023)Dao, Phung, Nguyen, and Tran]{dao2023flow}
Quan Dao, Hao Phung, Binh Nguyen, and Anh Tran.
\newblock Flow matching in latent space.
\newblock \emph{arXiv preprint arXiv:2307.08698}, 2023.

\bibitem[Dao et~al.(2024{\natexlab{a}})Dao, Phung, Dao, Metaxas, and Tran]{dao2024self}
Quan Dao, Hao Phung, Trung Dao, Dimitris Metaxas, and Anh Tran.
\newblock Self-corrected flow distillation for consistent one-step and few-step text-to-image generation.
\newblock \emph{arXiv preprint arXiv:2412.16906}, 2024{\natexlab{a}}.

\bibitem[Dao et~al.(2024{\natexlab{b}})Dao, Ta, Pham, and Tran]{dao2024high}
Quan Dao, Binh Ta, Tung Pham, and Anh Tran.
\newblock A high-quality robust diffusion framework for corrupted dataset.
\newblock In \emph{European Conference on Computer Vision}, pp.\  107--123. Springer, 2024{\natexlab{b}}.

\bibitem[Dhariwal \& Nichol(2021)Dhariwal and Nichol]{dhariwal2021diffusion}
Prafulla Dhariwal and Alexander Nichol.
\newblock Diffusion models beat gans on image synthesis.
\newblock \emph{Advances in neural information processing systems}, 34:\penalty0 8780--8794, 2021.

\bibitem[Geman \& Geman(1986)Geman and Geman]{geman1986bayesian}
Donald Geman and Stuart Geman.
\newblock Bayesian image analysis.
\newblock In \emph{Disordered systems and biological organization}, pp.\  301--319. Springer, 1986.

\bibitem[Geng et~al.(2024)Geng, Pokle, Luo, Lin, and Kolter]{geng2024consistency}
Zhengyang Geng, Ashwini Pokle, William Luo, Justin Lin, and J~Zico Kolter.
\newblock Consistency models made easy.
\newblock \emph{arXiv preprint arXiv:2406.14548}, 2024.

\bibitem[Goodfellow et~al.(2014)Goodfellow, Pouget-Abadie, Mirza, Xu, Warde-Farley, Ozair, Courville, and Bengio]{goodfellow2014generative}
Ian Goodfellow, Jean Pouget-Abadie, Mehdi Mirza, Bing Xu, David Warde-Farley, Sherjil Ozair, Aaron Courville, and Yoshua Bengio.
\newblock Generative adversarial nets.
\newblock \emph{Advances in neural information processing systems}, 27, 2014.

\bibitem[Gu et~al.(2022)Gu, Chen, Bao, Wen, Zhang, Chen, Yuan, and Guo]{gu2022vector}
Shuyang Gu, Dong Chen, Jianmin Bao, Fang Wen, Bo~Zhang, Dongdong Chen, Lu~Yuan, and Baining Guo.
\newblock Vector quantized diffusion model for text-to-image synthesis.
\newblock In \emph{Proceedings of the IEEE/CVF conference on computer vision and pattern recognition}, pp.\  10696--10706, 2022.

\bibitem[Han et~al.(2024)Han, Wen, Chen, Zhang, Song, Ren, Gao, Stathopoulos, He, Chen, et~al.]{han2024proxedit}
Ligong Han, Song Wen, Qi~Chen, Zhixing Zhang, Kunpeng Song, Mengwei Ren, Ruijiang Gao, Anastasis Stathopoulos, Xiaoxiao He, Yuxiao Chen, et~al.
\newblock Proxedit: Improving tuning-free real image editing with proximal guidance.
\newblock In \emph{Proceedings of the IEEE/CVF Winter Conference on Applications of Computer Vision}, pp.\  4291--4301, 2024.

\bibitem[He et~al.(2024)He, Han, Dao, Wen, Bai, Liu, Zhang, Min, Juefei-Xu, Tan, et~al.]{he2024dice}
Xiaoxiao He, Ligong Han, Quan Dao, Song Wen, Minhao Bai, Di~Liu, Han Zhang, Martin~Renqiang Min, Felix Juefei-Xu, Chaowei Tan, et~al.
\newblock Dice: Discrete inversion enabling controllable editing for multinomial diffusion and masked generative models.
\newblock \emph{arXiv preprint arXiv:2410.08207}, 2024.

\bibitem[Heek et~al.(2024)Heek, Hoogeboom, and Salimans]{heek2024multistep}
Jonathan Heek, Emiel Hoogeboom, and Tim Salimans.
\newblock Multistep consistency models.
\newblock \emph{arXiv preprint arXiv:2403.06807}, 2024.

\bibitem[Ho et~al.(2020)Ho, Jain, and Abbeel]{ho2020denoising}
Jonathan Ho, Ajay Jain, and Pieter Abbeel.
\newblock Denoising diffusion probabilistic models.
\newblock \emph{Advances in neural information processing systems}, 33:\penalty0 6840--6851, 2020.

\bibitem[Huang et~al.(2018)Huang, li, He, Sun, and Tan]{celeba}
Huaibo Huang, zhihang li, Ran He, Zhenan Sun, and Tieniu Tan.
\newblock Introvae: Introspective variational autoencoders for photographic image synthesis.
\newblock In S.~Bengio, H.~Wallach, H.~Larochelle, K.~Grauman, N.~Cesa-Bianchi, and R.~Garnett (eds.), \emph{Advances in Neural Information Processing Systems}, volume~31. Curran Associates, Inc., 2018.
\newblock URL \url{https://proceedings.neurips.cc/paper_files/paper/2018/file/093f65e080a295f8076b1c5722a46aa2-Paper.pdf}.

\bibitem[Huberman-Spiegelglas et~al.(2024)Huberman-Spiegelglas, Kulikov, and Michaeli]{huberman2024edit}
Inbar Huberman-Spiegelglas, Vladimir Kulikov, and Tomer Michaeli.
\newblock An edit friendly ddpm noise space: Inversion and manipulations.
\newblock In \emph{Proceedings of the IEEE/CVF Conference on Computer Vision and Pattern Recognition}, pp.\  12469--12478, 2024.

\bibitem[Kang et~al.(2024)Kang, Zhang, Barnes, Paris, Kwak, Park, Shechtman, Zhu, and Park]{kang2024diffusion2gan}
Minguk Kang, Richard Zhang, Connelly Barnes, Sylvain Paris, Suha Kwak, Jaesik Park, Eli Shechtman, Jun-Yan Zhu, and Taesung Park.
\newblock {Distilling Diffusion Models into Conditional GANs}.
\newblock In \emph{European Conference on Computer Vision (ECCV)}, 2024.

\bibitem[Karras et~al.(2019)Karras, Laine, and Aila]{karras2019style}
Tero Karras, Samuli Laine, and Timo Aila.
\newblock A style-based generator architecture for generative adversarial networks.
\newblock In \emph{Proceedings of the IEEE/CVF conference on computer vision and pattern recognition}, pp.\  4401--4410, 2019.

\bibitem[Karras et~al.(2022)Karras, Aittala, Aila, and Laine]{Karras2022edm}
Tero Karras, Miika Aittala, Timo Aila, and Samuli Laine.
\newblock Elucidating the design space of diffusion-based generative models.
\newblock In \emph{Proc. NeurIPS}, 2022.

\bibitem[Kim et~al.(2021)Kim, Shin, Song, Kang, and Moon]{kim2021soft}
Dongjun Kim, Seungjae Shin, Kyungwoo Song, Wanmo Kang, and Il-Chul Moon.
\newblock Soft truncation: A universal training technique of score-based diffusion model for high precision score estimation.
\newblock \emph{arXiv preprint arXiv:2106.05527}, 2021.

\bibitem[Kim et~al.(2023)Kim, Lai, Liao, Murata, Takida, Uesaka, He, Mitsufuji, and Ermon]{kim2023consistency}
Dongjun Kim, Chieh-Hsin Lai, Wei-Hsiang Liao, Naoki Murata, Yuhta Takida, Toshimitsu Uesaka, Yutong He, Yuki Mitsufuji, and Stefano Ermon.
\newblock Consistency trajectory models: Learning probability flow ode trajectory of diffusion.
\newblock \emph{arXiv preprint arXiv:2310.02279}, 2023.

\bibitem[Kong et~al.(2023)Kong, Duan, Sun, Cheng, Xu, Shen, Zhu, Shi, and Xu]{kong2023act}
Fei Kong, Jinhao Duan, Lichao Sun, Hao Cheng, Renjing Xu, Hengtao Shen, Xiaofeng Zhu, Xiaoshuang Shi, and Kaidi Xu.
\newblock Act: Adversarial consistency models.
\newblock \emph{arXiv preprint arXiv:2311.14097}, 2023.

\bibitem[Kumari et~al.(2023)Kumari, Zhang, Zhang, Shechtman, and Zhu]{kumari2023multi}
Nupur Kumari, Bingliang Zhang, Richard Zhang, Eli Shechtman, and Jun-Yan Zhu.
\newblock Multi-concept customization of text-to-image diffusion.
\newblock In \emph{Proceedings of the IEEE/CVF Conference on Computer Vision and Pattern Recognition}, pp.\  1931--1941, 2023.

\bibitem[Kynk{\"a}{\"a}nniemi et~al.(2019)Kynk{\"a}{\"a}nniemi, Karras, Laine, Lehtinen, and Aila]{kynkaanniemi2019improved}
Tuomas Kynk{\"a}{\"a}nniemi, Tero Karras, Samuli Laine, Jaakko Lehtinen, and Timo Aila.
\newblock Improved precision and recall metric for assessing generative models.
\newblock \emph{Advances in Neural Information Processing Systems}, 32, 2019.

\bibitem[Lee \& He(2019)Lee and He]{lee2019target}
Donghwan Lee and Niao He.
\newblock Target-based temporal-difference learning.
\newblock In \emph{International Conference on Machine Learning}, pp.\  3713--3722. PMLR, 2019.

\bibitem[Lee et~al.(2023)Lee, Kim, and Ye]{lee2023minimizing}
Sangyun Lee, Beomsu Kim, and Jong~Chul Ye.
\newblock Minimizing trajectory curvature of ode-based generative models.
\newblock \emph{arXiv preprint arXiv:2301.12003}, 2023.

\bibitem[Luo et~al.(2023)Luo, Tan, Huang, Li, and Zhao]{luo2023latent}
Simian Luo, Yiqin Tan, Longbo Huang, Jian Li, and Hang Zhao.
\newblock Latent consistency models: Synthesizing high-resolution images with few-step inference, 2023.

\bibitem[Meng et~al.(2021)Meng, He, Song, Song, Wu, Zhu, and Ermon]{meng2021sdedit}
Chenlin Meng, Yutong He, Yang Song, Jiaming Song, Jiajun Wu, Jun-Yan Zhu, and Stefano Ermon.
\newblock Sdedit: Guided image synthesis and editing with stochastic differential equations.
\newblock \emph{arXiv preprint arXiv:2108.01073}, 2021.

\bibitem[Meng et~al.(2023)Meng, Rombach, Gao, Kingma, Ermon, Ho, and Salimans]{meng2023distillation}
Chenlin Meng, Robin Rombach, Ruiqi Gao, Diederik Kingma, Stefano Ermon, Jonathan Ho, and Tim Salimans.
\newblock On distillation of guided diffusion models.
\newblock In \emph{Proceedings of the IEEE/CVF Conference on Computer Vision and Pattern Recognition}, pp.\  14297--14306, 2023.

\bibitem[Naeem et~al.(2020)Naeem, Oh, Uh, Choi, and Yoo]{fid}
Muhammad~Ferjad Naeem, Seong~Joon Oh, Youngjung Uh, Yunjey Choi, and Jaejun Yoo.
\newblock Reliable fidelity and diversity metrics for generative models.
\newblock \emph{ArXiv}, abs/2002.09797, 2020.
\newblock URL \url{https://api.semanticscholar.org/CorpusID:211259260}.

\bibitem[Park et~al.(2024)Park, Kim, Lee, and Kim]{park2024ddmi}
Dogyun Park, Sihyeon Kim, Sojin Lee, and Hyunwoo~J Kim.
\newblock Ddmi: Domain-agnostic latent diffusion models for synthesizing high-quality implicit neural representations.
\newblock \emph{arXiv preprint arXiv:2401.12517}, 2024.

\bibitem[Phung et~al.(2023)Phung, Dao, and Tran]{phung2023wavediff}
Hao Phung, Quan Dao, and Anh Tran.
\newblock Wavelet diffusion models are fast and scalable image generators.
\newblock In \emph{Proceedings of the IEEE/CVF Conference on Computer Vision and Pattern Recognition (CVPR)}, pp.\  10199--10208, June 2023.

\bibitem[Phung et~al.(2024)Phung, Dao, Dao, Phan, Metaxas, and Tran]{phung2024dimsum}
Hao Phung, Quan Dao, Trung Dao, Hoang Phan, Dimitris Metaxas, and Anh Tran.
\newblock Dimsum: Diffusion mamba--a scalable and unified spatial-frequency method for image generation.
\newblock \emph{arXiv preprint arXiv:2411.04168}, 2024.

\bibitem[Pooladian et~al.(2023)Pooladian, Ben-Hamu, Domingo-Enrich, Amos, Lipman, and Chen]{pooladian2023multisample}
Aram-Alexandre Pooladian, Heli Ben-Hamu, Carles Domingo-Enrich, Brandon Amos, Yaron Lipman, and Ricky~TQ Chen.
\newblock Multisample flow matching: Straightening flows with minibatch couplings.
\newblock \emph{arXiv preprint arXiv:2304.14772}, 2023.

\bibitem[Poole et~al.(2022)Poole, Jain, Barron, and Mildenhall]{poole2022dreamfusion}
Ben Poole, Ajay Jain, Jonathan~T. Barron, and Ben Mildenhall.
\newblock Dreamfusion: Text-to-3d using 2d diffusion.
\newblock \emph{arXiv}, 2022.

\bibitem[Ren et~al.(2024)Ren, Xia, Lu, Zhang, Wu, Xie, Wang, and Xiao]{ren2024hyper}
Yuxi Ren, Xin Xia, Yanzuo Lu, Jiacheng Zhang, Jie Wu, Pan Xie, Xing Wang, and Xuefeng Xiao.
\newblock Hyper-sd: Trajectory segmented consistency model for efficient image synthesis.
\newblock \emph{arXiv preprint arXiv:2404.13686}, 2024.

\bibitem[Rombach et~al.(2021)Rombach, Blattmann, Lorenz, Esser, and Ommer]{rombach2021highresolution}
Robin Rombach, Andreas Blattmann, Dominik Lorenz, Patrick Esser, and Björn Ommer.
\newblock High-resolution image synthesis with latent diffusion models, 2021.

\bibitem[Ruiz et~al.(2022)Ruiz, Li, Jampani, Pritch, Rubinstein, and Aberman]{ruiz2022dreambooth}
Nataniel Ruiz, Yuanzhen Li, Varun Jampani, Yael Pritch, Michael Rubinstein, and Kfir Aberman.
\newblock Dreambooth: Fine tuning text-to-image diffusion models for subject-driven generation.
\newblock \emph{arXiv preprint}, 2022.

\bibitem[Sauer et~al.(2023)Sauer, Lorenz, Blattmann, and Rombach]{sauer2023adversarial}
Axel Sauer, Dominik Lorenz, Andreas Blattmann, and Robin Rombach.
\newblock Adversarial diffusion distillation.
\newblock \emph{arXiv preprint arXiv:2311.17042}, 2023.

\bibitem[Sohl-Dickstein et~al.(2015)Sohl-Dickstein, Weiss, Maheswaranathan, and Ganguli]{sohl2015deep}
Jascha Sohl-Dickstein, Eric Weiss, Niru Maheswaranathan, and Surya Ganguli.
\newblock Deep unsupervised learning using nonequilibrium thermodynamics.
\newblock In \emph{International conference on machine learning}, pp.\  2256--2265. PMLR, 2015.

\bibitem[Song \& Dhariwal(2023)Song and Dhariwal]{song2023improved}
Yang Song and Prafulla Dhariwal.
\newblock Improved techniques for training consistency models.
\newblock \emph{arXiv preprint arXiv:2310.14189}, 2023.

\bibitem[Song \& Ermon(2019)Song and Ermon]{song2019generative}
Yang Song and Stefano Ermon.
\newblock Generative modeling by estimating gradients of the data distribution.
\newblock \emph{Advances in neural information processing systems}, 32, 2019.

\bibitem[Song et~al.(2020)Song, Sohl-Dickstein, Kingma, Kumar, Ermon, and Poole]{song2020score}
Yang Song, Jascha Sohl-Dickstein, Diederik~P Kingma, Abhishek Kumar, Stefano Ermon, and Ben Poole.
\newblock Score-based generative modeling through stochastic differential equations.
\newblock \emph{arXiv preprint arXiv:2011.13456}, 2020.

\bibitem[Song et~al.(2023)Song, Dhariwal, Chen, and Sutskever]{song2023consistency}
Yang Song, Prafulla Dhariwal, Mark Chen, and Ilya Sutskever.
\newblock Consistency models.
\newblock \emph{arXiv preprint arXiv:2303.01469}, 2023.

\bibitem[Teng et~al.(2023)Teng, Zheng, Ding, Hong, Wangni, Yang, and Tang]{teng2023relay}
Jiayan Teng, Wendi Zheng, Ming Ding, Wenyi Hong, Jianqiao Wangni, Zhuoyi Yang, and Jie Tang.
\newblock Relay diffusion: Unifying diffusion process across resolutions for image synthesis.
\newblock \emph{arXiv preprint arXiv:2309.03350}, 2023.

\bibitem[Tong et~al.(2023)Tong, Malkin, Huguet, Zhang, Rector-Brooks, Fatras, Wolf, and Bengio]{tong2023improving}
Alexander Tong, Nikolay Malkin, Guillaume Huguet, Yanlei Zhang, Jarrid Rector-Brooks, Kilian Fatras, Guy Wolf, and Yoshua Bengio.
\newblock Improving and generalizing flow-based generative models with minibatch optimal transport.
\newblock \emph{arXiv preprint arXiv:2302.00482}, 2023.

\bibitem[Vahdat et~al.(2021)Vahdat, Kreis, and Kautz]{vahdat2021score}
Arash Vahdat, Karsten Kreis, and Jan Kautz.
\newblock Score-based generative modeling in latent space.
\newblock \emph{Advances in neural information processing systems}, 34:\penalty0 11287--11302, 2021.

\bibitem[Van~Le et~al.(2023)Van~Le, Phung, Nguyen, Dao, Tran, and Tran]{van2023anti}
Thanh Van~Le, Hao Phung, Thuan~Hoang Nguyen, Quan Dao, Ngoc~N Tran, and Anh Tran.
\newblock Anti-dreambooth: Protecting users from personalized text-to-image synthesis.
\newblock In \emph{Proceedings of the IEEE/CVF International Conference on Computer Vision}, pp.\  2116--2127, 2023.

\bibitem[Wang et~al.(2024)Wang, Lu, Wang, Bao, Li, Su, and Zhu]{wang2024prolificdreamer}
Zhengyi Wang, Cheng Lu, Yikai Wang, Fan Bao, Chongxuan Li, Hang Su, and Jun Zhu.
\newblock Prolificdreamer: High-fidelity and diverse text-to-3d generation with variational score distillation.
\newblock \emph{Advances in Neural Information Processing Systems}, 36, 2024.

\bibitem[Wei et~al.(2022)Wei, Zhang, Zhang, Gong, Zhang, Zhang, Yu, and Liu]{wei2022outlier}
Xiuying Wei, Yunchen Zhang, Xiangguo Zhang, Ruihao Gong, Shanghang Zhang, Qi~Zhang, Fengwei Yu, and Xianglong Liu.
\newblock Outlier suppression: Pushing the limit of low-bit transformer language models.
\newblock \emph{Advances in Neural Information Processing Systems}, 35:\penalty0 17402--17414, 2022.

\bibitem[Wu \& la~Torre(2023)Wu and la~Torre]{cyclediffusion}
Chen~Henry Wu and Fernando~De la~Torre.
\newblock A latent space of stochastic diffusion models for zero-shot image editing and guidance.
\newblock In \emph{ICCV}, 2023.

\bibitem[Xiao et~al.(2021)Xiao, Kreis, and Vahdat]{xiao2021tackling}
Zhisheng Xiao, Karsten Kreis, and Arash Vahdat.
\newblock Tackling the generative learning trilemma with denoising diffusion gans.
\newblock \emph{arXiv preprint arXiv:2112.07804}, 2021.

\bibitem[Yin et~al.(2024)Yin, Gharbi, Zhang, Shechtman, Durand, Freeman, and Park]{yin2024one}
Tianwei Yin, Micha{\"e}l Gharbi, Richard Zhang, Eli Shechtman, Fredo Durand, William~T Freeman, and Taesung Park.
\newblock One-step diffusion with distribution matching distillation.
\newblock In \emph{Proceedings of the IEEE/CVF Conference on Computer Vision and Pattern Recognition}, pp.\  6613--6623, 2024.

\bibitem[Yu et~al.(2015)Yu, Zhang, Song, Seff, and Xiao]{lsun}
Fisher Yu, Yinda Zhang, Shuran Song, Ari Seff, and Jianxiong Xiao.
\newblock Lsun: Construction of a large-scale image dataset using deep learning with humans in the loop.
\newblock \emph{ArXiv}, abs/1506.03365, 2015.
\newblock URL \url{https://api.semanticscholar.org/CorpusID:8317437}.

\bibitem[Zhang et~al.(2023{\natexlab{a}})Zhang, Zhou, Lu, Guo, Wang, Shen, and Qu]{zhang2023emergence}
Huijie Zhang, Jinfan Zhou, Yifu Lu, Minzhe Guo, Peng Wang, Liyue Shen, and Qing Qu.
\newblock The emergence of reproducibility and consistency in diffusion models.
\newblock In \emph{Forty-first International Conference on Machine Learning}, 2023{\natexlab{a}}.

\bibitem[Zhang et~al.(2023{\natexlab{b}})Zhang, Rao, and Agrawala]{zhang2023adding}
Lvmin Zhang, Anyi Rao, and Maneesh Agrawala.
\newblock Adding conditional control to text-to-image diffusion models, 2023{\natexlab{b}}.

\bibitem[Zhangli et~al.(2024)Zhangli, Jiang, Liu, Yu, Dai, Ramchandani, Pang, Metaxas, and Krishnan]{zhangli2024layout}
Qilong Zhangli, Jindong Jiang, Di~Liu, Licheng Yu, Xiaoliang Dai, Ankit Ramchandani, Guan Pang, Dimitris~N Metaxas, and Praveen Krishnan.
\newblock Layout-agnostic scene text image synthesis with diffusion models.
\newblock In \emph{2024 IEEE/CVF Conference on Computer Vision and Pattern Recognition (CVPR)}, pp.\  7496--7506. IEEE Computer Society, 2024.

\bibitem[Zheng et~al.(2024)Zheng, Hu, Fan, Wang, Ding, Tao, and Cham]{zheng2024trajectory}
Jianbin Zheng, Minghui Hu, Zhongyi Fan, Chaoyue Wang, Changxing Ding, Dacheng Tao, and Tat-Jen Cham.
\newblock Trajectory consistency distillation.
\newblock \emph{arXiv preprint arXiv:2402.19159}, 2024.

\end{thebibliography}
\bibliographystyle{iclr2025_conference}

\newpage
\appendix
\section{Appendix}
\begin{figure}[h]
\centering
    \includegraphics[width=0.6\textwidth]{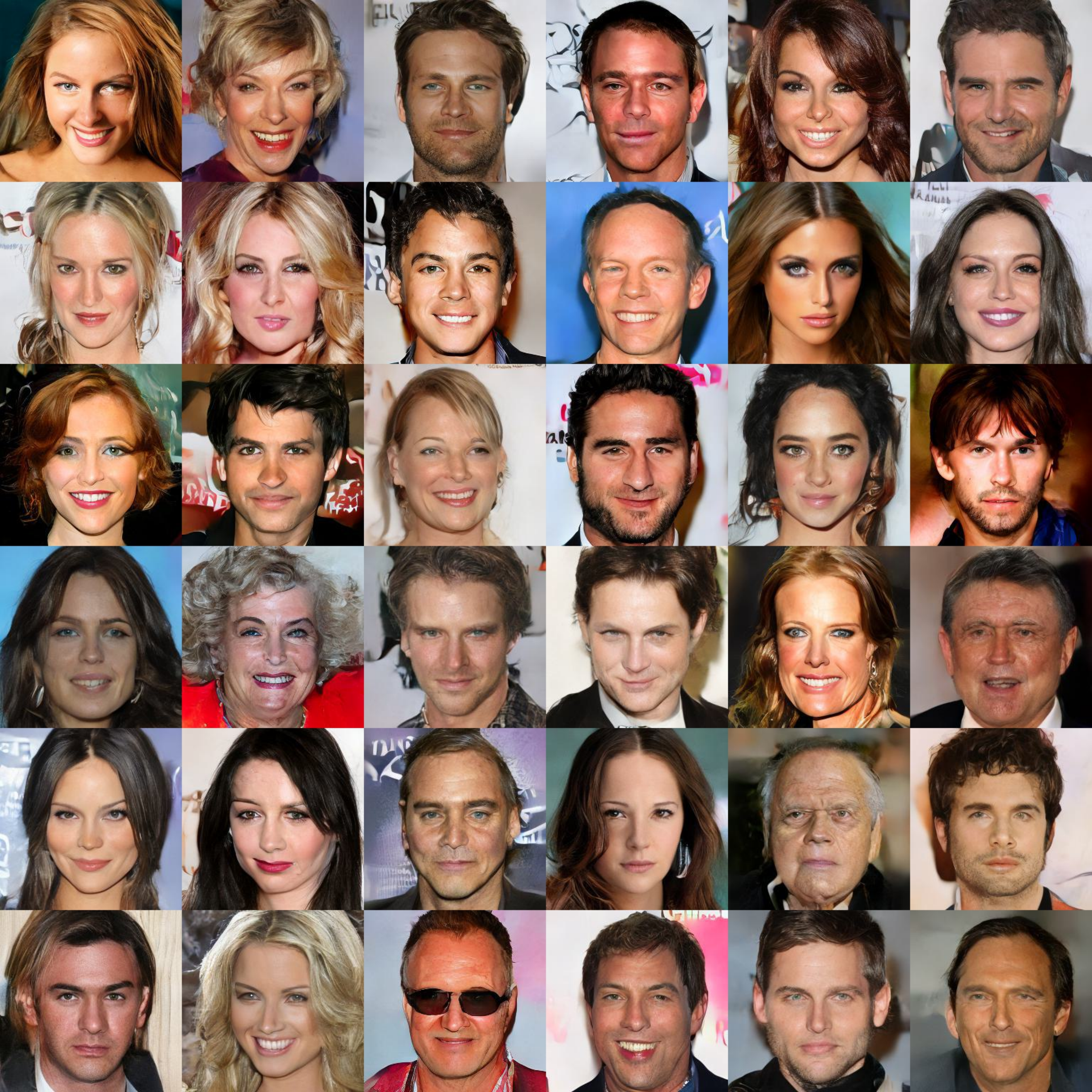}
    \caption{One-step samples on CelebA-HQ $256 \times 256$}
    \label{fig:appendix:celeba_onestep}
\end{figure}

\begin{figure}[h]
\centering
    \includegraphics[width=0.6\textwidth]{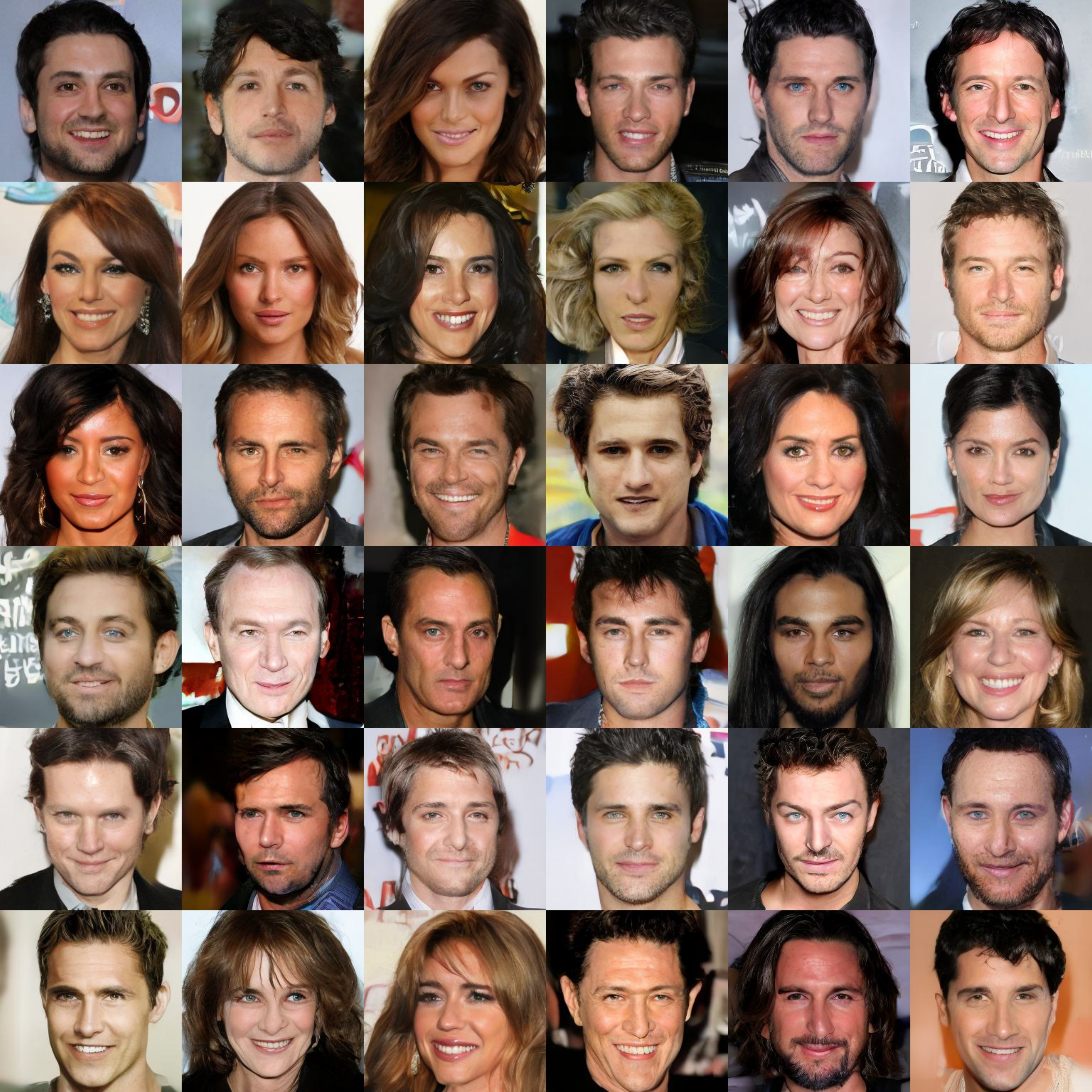}
    \caption{Two-step samples on CelebA-HQ $256 \times 256$}
    \label{fig:appendix:celeba_twostep}
\end{figure}

\begin{figure}[h]
\centering
    \includegraphics[width=0.6\textwidth]{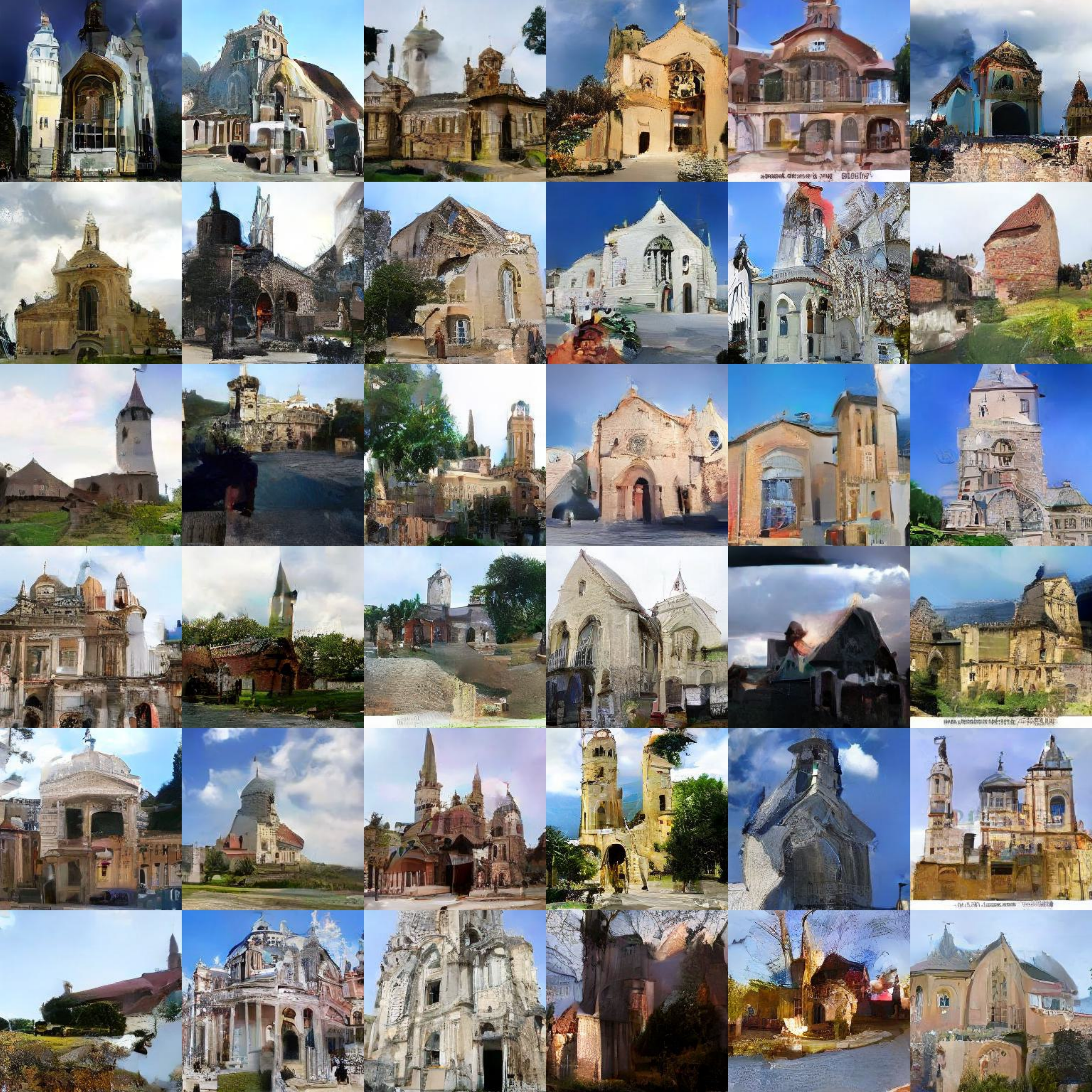}
    \caption{One-step samples on LSUN Church $256 \times 256$}
    \label{fig:appendix:lsun_onestep}
\end{figure}

\begin{figure}[h]
\centering
    \includegraphics[width=0.6\textwidth]{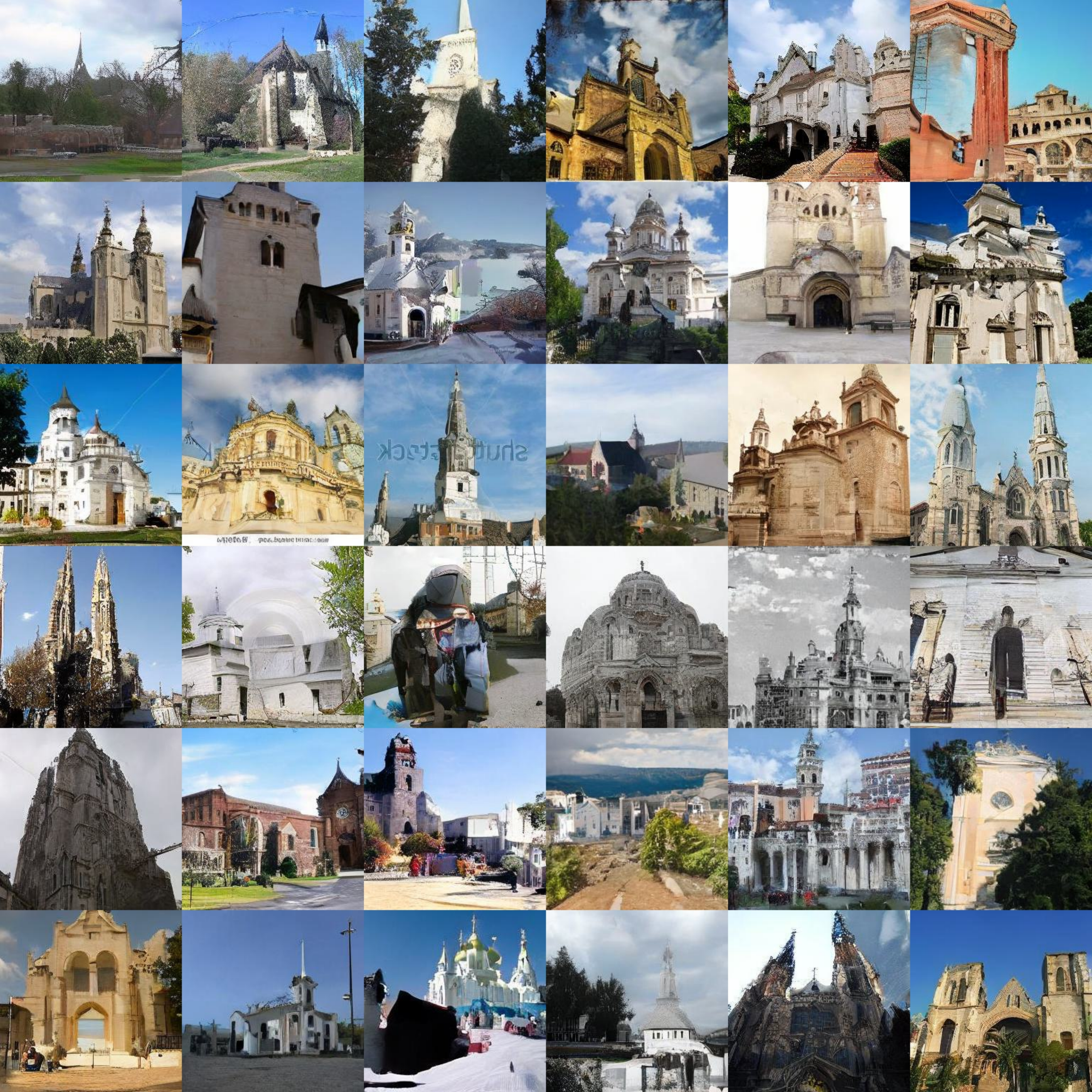}
    \caption{Two-step samples on LSUN Church $256 \times 256$}
    \label{fig:appendix:lsun_twostep}
\end{figure}

\begin{figure}[h]
\centering
    \includegraphics[width=0.6\textwidth]{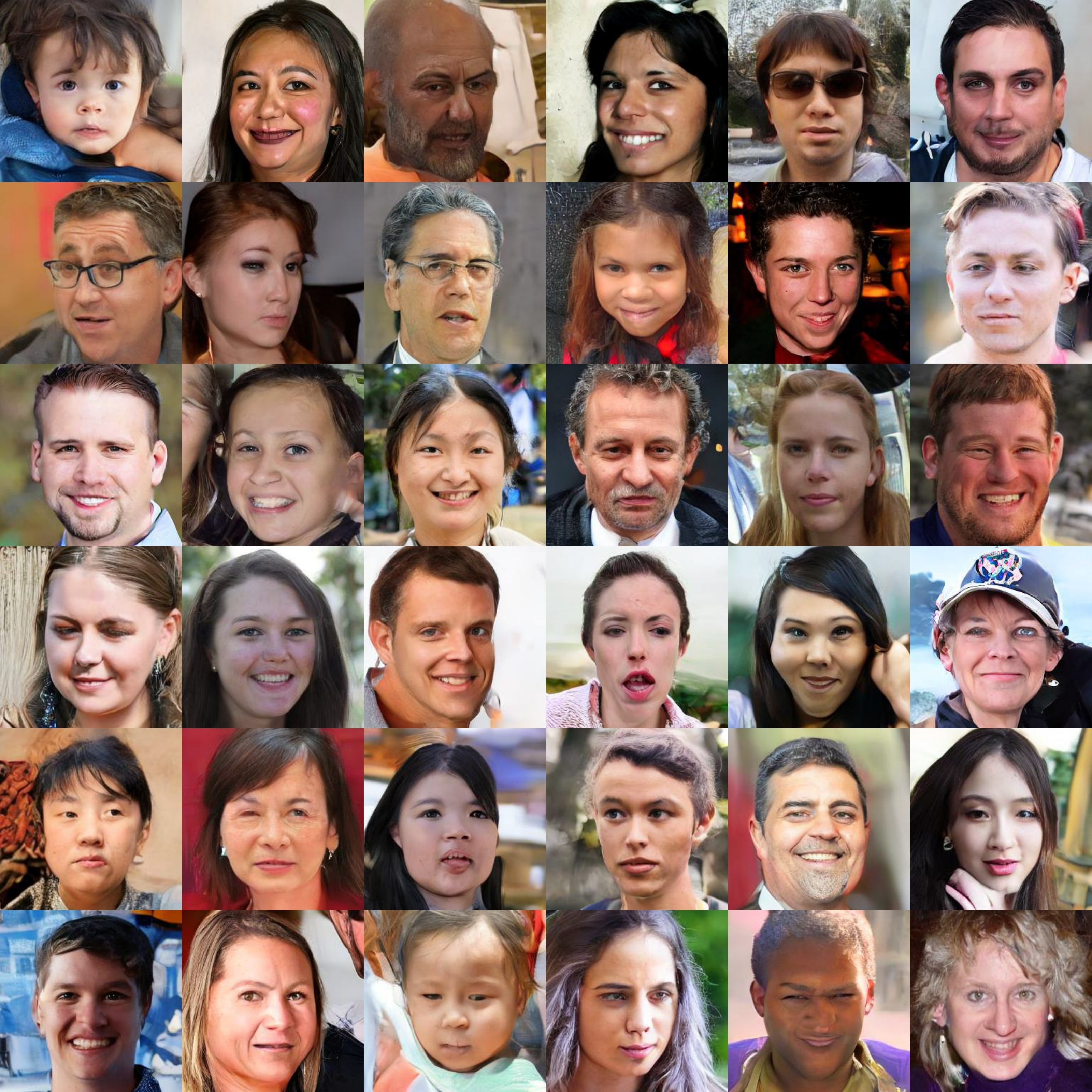}
    \caption{One-step samples on FFHQ $256 \times 256$}
    \label{fig:appendix:ffhq_onestep}
\end{figure}

\begin{figure}[h]
\centering
    \includegraphics[width=0.6\textwidth]{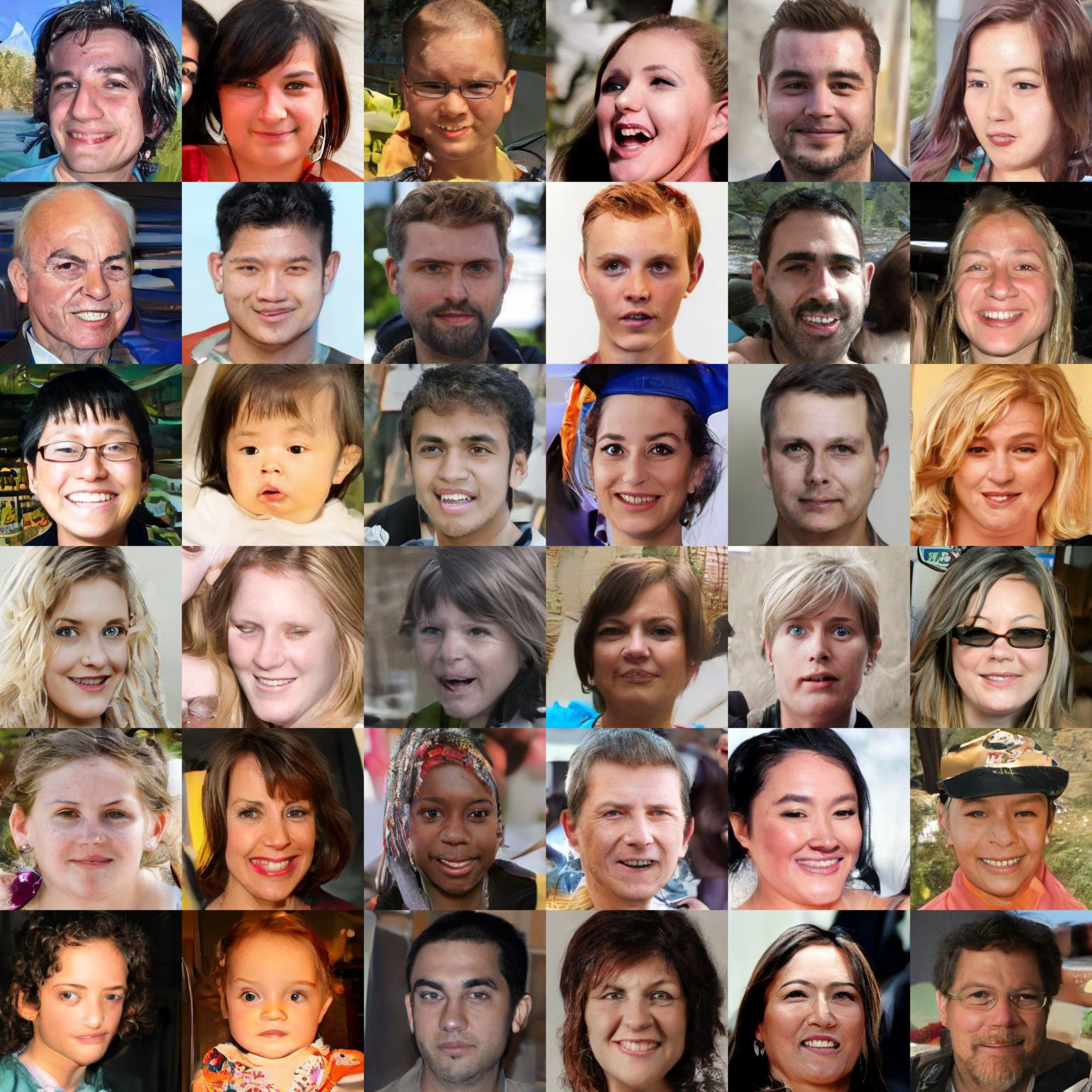}
    \caption{Two-step samples on FFHQ $256 \times 256$}
    \label{fig:appendix:ffhq_twostep}
\end{figure}

\begin{figure}[h]
\centering
    \includegraphics[width=0.6\textwidth]{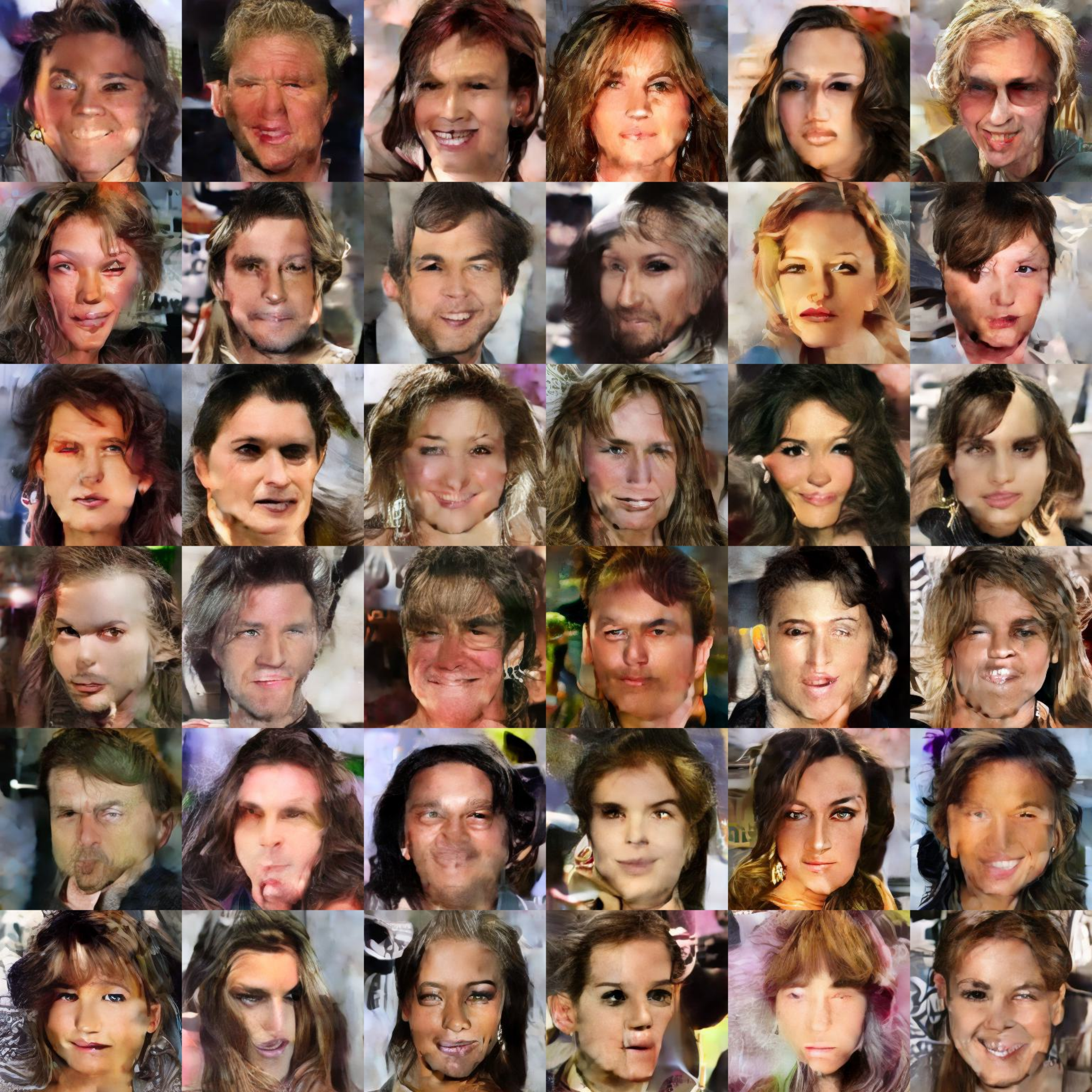}
    \caption{One-step samples on CelebA-HQ $256 \times 256$ (L2 loss)}
    \label{fig:appendix:celeba_onestep_ilct_l2}
\end{figure}

\begin{figure}[h]
\centering
    \includegraphics[width=0.6\textwidth]{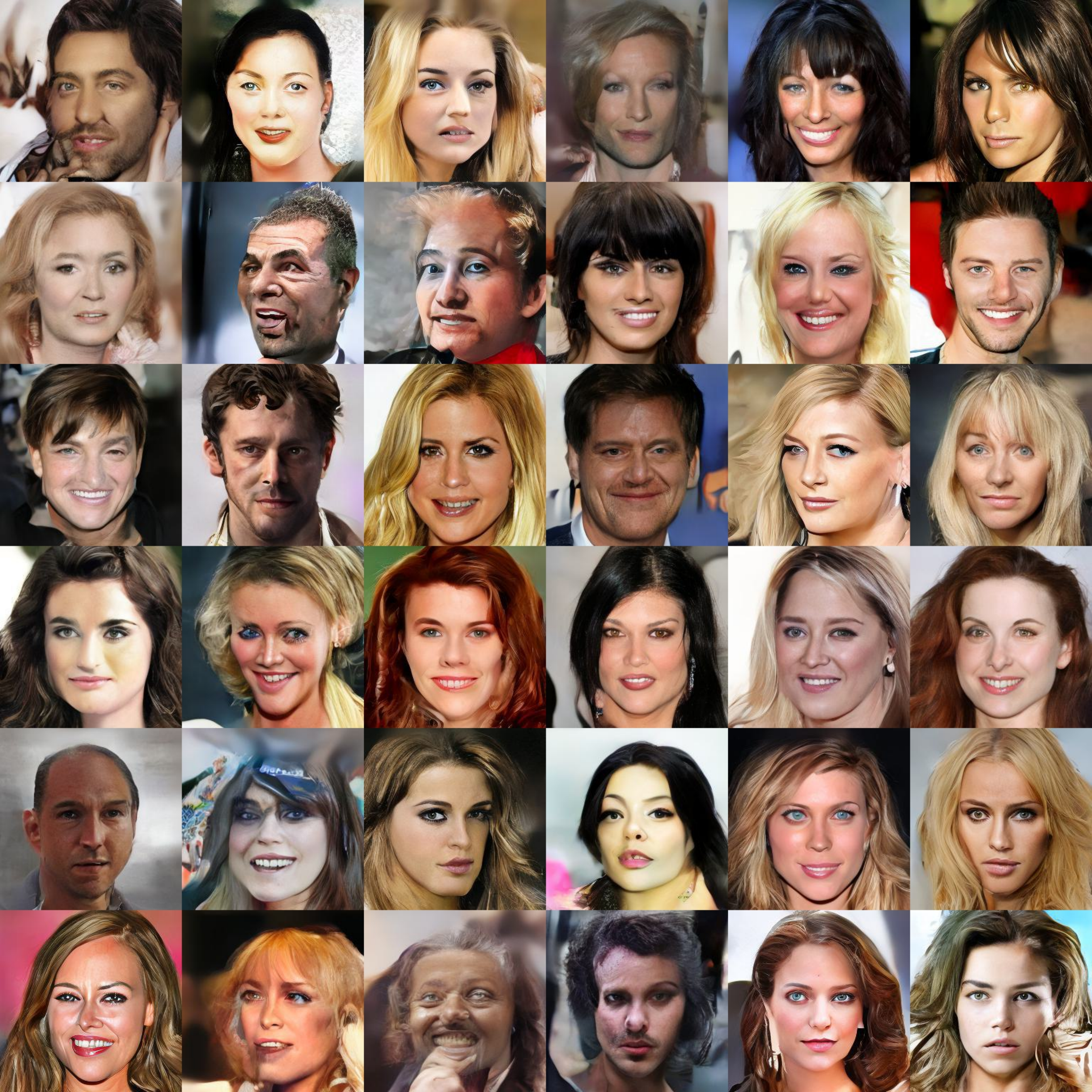}
    \caption{One-step samples on CelebA-HQ $256 \times 256$ (E-LatentLPIPS loss)}
    \label{fig:appendix:celeba_onestep_ilct_elatentlpips}
\end{figure}

We provide additional uncurated samples of our models for three datasets: CelebaA-HQ (\ref{fig:appendix:celeba_onestep}, \ref{fig:appendix:celeba_twostep}), LSUN Church (\ref{fig:appendix:lsun_onestep}, \ref{fig:appendix:lsun_twostep}), and FFHQ (\ref{fig:appendix:ffhq_onestep}, \ref{fig:appendix:ffhq_twostep}). We also provide additional uncurated samples of our models on CelebaA-HQ trained with L2 loss (\ref{fig:appendix:celeba_onestep_ilct_l2}) and E-LatentLPIPS loss (\ref{fig:appendix:celeba_onestep_ilct_elatentlpips}).

\end{document}